# Ordered Landmarks in Planning

**Jörg Hoffmann** HOFFMANN@MPI-SB.MPG.DE
*Max-Planck-Institut für Informatik,*
*Saarbrücken, Germany*
**Julie Porteous** JULIE.PORTEOUS@CIS.STRATH.AC.UK
*Department of Computer and Information Sciences,*
*The University of Strathclyde,*
*Glasgow, UK*
**Laura Sebastia** LSTARIN@DSIC.UPV.ES
*Dpto. Sist. Informáticos y Computación,*
*Universidad Politécnica de Valencia,*
*Valencia, Spain*

## Abstract

Many known planning tasks have inherent constraints concerning the best order in which to achieve the goals. A number of research efforts have been made to detect such constraints and to use them for guiding search, in the hope of speeding up the planning process.

We go beyond the previous approaches by considering ordering constraints not only over the (top-level) goals, but also over the sub-goals that will necessarily arise during planning. Landmarks are facts that must be true at some point in every valid solution plan. We extend Koehler and Hoffmann's definition of *reasonable* orders between top level goals to the more general case of landmarks. We show how landmarks can be found, how their reasonable orders can be approximated, and how this information can be used to decompose a given planning task into several smaller sub-tasks. Our methodology is completely domain- and planner-independent. The implementation demonstrates that the approach can yield significant runtime performance improvements when used as a control loop around state-of-the-art sub-optimal planning systems, as exemplified by FF and LPG.

## 1. Introduction

Given the inherent complexity of the general planning problem it is clearly important to develop good heuristic strategies for both managing and navigating the search space involved in solving a particular planning instance. One way in which search can be informed is by providing hints concerning the order in which planning goals should be addressed. This can make a significant difference to search efficiency by helping to focus the planner on a progressive path towards a solution. Work in this area includes that of Koehler and Hoffmann (2000). They introduce the notion of *reasonable orders* which states that a pair of goals $A$ and $B$ can be ordered so that $B$ is achieved before $A$ if it isn't possible to reach a state in which $A$ and $B$ are both true, from a state in which just $A$ is true, without having to temporarily destroy $A$. In such a situation it is reasonable to achieve $B$ before $A$ to avoid unnecessary effort.

The main idea behind the work discussed in this paper is to extend those previous ideas on orders by not only ordering the (top-level) goals, but also the sub-goals that will necessarily arise during planning, i.e., by also taking into account what we call the *landmarks.* The key feature of a landmark is that it *must* be true at some point on any solution path to the given planning task. Consider the *Blocksworld* task shown in Figure 1, which will be our illustrative example throughout the paper.

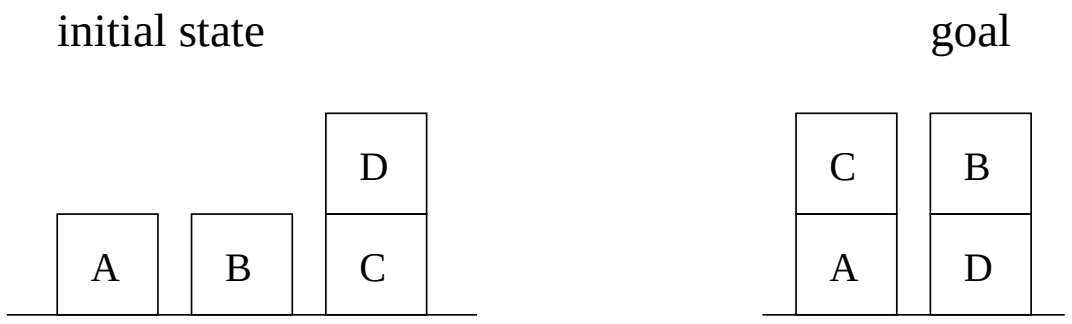

Figure 1: Example *Blocksworld* task.

For the reader who is weary of seeing toy examples like the one in Figure 1 in the literature, we remark that our techniques are *not* primarily motivated by this example. Our techniques are useful in much more complex situations. We use the depicted toy example only for easy demonstration of some of the important points. In the example, *clear(C)* is a landmark because it will need to be achieved in any solution plan. Immediately stacking B on D from the initial state will achieve one of the top level goals of the task but it will result in wasted effort if *clear(C)* is not achieved first. To order *clear(C)* before *on(B D)* is, however, *not* reasonable in terms of Koehler and Hoffmann's definition. First, *clear(C)* is not a top level goal so it is not considered by Koehler and Hoffmann's techniques. Second, there are states where B is on D and from which *clear(C)* can be achieved without unstacking B from D again (compare the definition of reasonable orders given above). But reaching such a state requires unstacking D from C, and thus achieving *clear(C)*, in the first place. This, together with the fact that *clear(C)* must be made true at some point, makes it sensible to order *clear(C)* before *on(B D)*.

We propose a natural extension of Koehler and Hoffmann's definitions to the more general case of landmarks (trivially, all top level goals are landmarks, too). We also revise parts of the original definition to better capture the intuitive meaning of a goal ordering. The extended and revised definitions capture, in particular, situations of the kind demonstrated with *clear(C)* $\leq$ *on(B D)* in the toy example above. We also introduce a new kind of ordering that often occurs between landmarks: $A$ can be ordered before $B$ if all valid solution plans make $A$ true before they make $B$ true. We call such orders *necessary.* Typically, a fact *is* a landmark because it is necessarily ordered before some other landmark. For example, *clear(C)* is necessarily ordered before *holding(C)*, and *holding(C)* is necessarily ordered before the top level goal *on(C A)*, in the above *Blocksworld* example.

Deciding if a fact is a landmark, and deciding about our ordering relations, is PSPACE-complete. We describe pre-processing techniques that extract landmarks, and that approximate necessary orders between them. We introduce sufficient criteria for the existence of reasonable orders between landmarks. The criteria are based on necessary orders, and inconsistencies between facts.[1] Using an inconsistency approximation technique from the

1. Two facts are inconsistent if they are not true together in any reachable world state.

literature, we approximate reasonable orders based on our sufficient criteria. After these pre-processes have terminated, what we get is a directed graph where the nodes are the found landmarks, and the edges are the orders found between them. We call this graph the *landmark generation graph*, short LGG. This graph may contain cycles because for some of our orders there is no guarantee that there is a plan, or even an action sequence, that obeys them.[2] Our method for structuring the search for a plan can not handle cycles in the LGG, so we remove cycles by removing edges incident upon them. We end up with a polytree structure.[3]

Once turned into a polytree, the LGG can be used to decompose the planning task into small chunks. We propose a method that does not depend on any particular planning framework. The landmarks provide a search control loop that can be used around any base planner that is capable of dealing with STRIPS input. The search control does not preserve optimality so there is not much point in using it around optimal planners such as Graphplan (Blum & Furst, 1997) and its relatives. Optimal planners are generally outperformed by sub-optimal planners anyway. It does make sense, however, to use the control in order to further improve the runtime performance of sub-optimal approaches to planning. To demonstrate this, we used the technique for control of two versions of FF (Hoffmann, 2000; Hoffmann & Nebel, 2001), and for control of LPG (Gerevini, Saetti, & Serina, 2003). We evaluated these planners across a range of 8 domains. We consistently obtain, sometimes dramatic, runtime improvements for the FF versions. We obtain runtime improvements for LPG in around half of the domains. The runtime improvement is, for all the planners, usually bought at the cost of slightly longer plans. But there are also some cases where the plans become shorter when using landmarks control.

The paper is organised as follows. Section 2 gives the basic notations. Section 3 defines what landmarks are, and in what relations between them we are interested. Exact computation of the relevant pieces of information is shown to be PSPACE-complete. Section 4 explains our approximation techniques, and Section 5 explains how we use landmarks to structure the search of an arbitrary base planner. Section 6 provides our empirical results in a range of domains. Section 7 closes the paper with a discussion of related work, of our contributions, and of future work. Most proofs are moved into Appendix A, and replaced in the text by proof sketches, to improve readability. Appendix B provides runtime distribution graphs as supplementary material to the tables provided in Section 6. Appendix C discusses some details regarding our experimental implementation of landmarks control around LPG.

## 2. Notations

We consider sequential planning in the propositional STRIPS (Fikes & Nilsson, 1971) framework. In the following, all sets are assumed to be finite. A *state* $s$ is a set of logical facts (atoms). An *action* $a$ is a triple $a = (pre(a), add(a), del(a))$ where $pre(a)$ are the action's *preconditions*, $add(a)$ is its *add list*, and $del(a)$ is its *delete list*, each a set of facts. The

2. Also, none of our ordering relations is transitive. We stick to the word "order" only because it is the most intuitive word for constraints on the relative points in time at which planning facts can or should be achieved.
3. Removing edges incident on cycles might, of course, throw away useful ordering information. Coming up with other methods to treat cycles, or with methods that can exploit the information contained in them, is an open research topic.

result of applying (the action sequence consisting of) a single action $a$ to a state $s$ is:

$$Result(s, \langle a \rangle) = \begin{cases} (s \cup add(a)) \setminus del(a) & pre(a) \subseteq s \\ \text{undefined} & \text{otherwise} \end{cases}$$

The result of applying a sequence of more than one action to a state is recursively defined as $Result(s, \langle a_1, \ldots, a_n \rangle) = Result(Result(s, \langle a_1, \ldots, a_{n-1} \rangle), \langle a_n \rangle)$. Applying an empty action sequence changes nothing, i.e., $Result(s, \langle\rangle) = s$. A planning *task* $(A, I, G)$ is a triple where $A$ is a set of actions, and $I$ (the initial state) and $G$ (the goals) are sets of facts (we use the word "task" rather than "problem" in order to avoid confusion with the complexity-theoretic notion of decision problems). A *plan*, or *solution*, for a task $(A, I, G)$ is an action sequence $P \in A^*$ such that $G \subseteq Result(I, P)$.

## 3. Ordered Landmarks: What They Are

In this section we introduce our framework. We define what landmarks are, and in what relations between them we are interested. We show that all the corresponding decision problems are PSPACE-complete. Section 3.1 introduces landmarks and necessary orders, Section 3.2 introduces reasonable orders, and Section 3.3 introduces *obedient* reasonable orders (orders that are reasonable if one has already committed to obey a given a-priori set of reasonable ordering constraints).

### 3.1 Landmarks, and Necessary Orders

Landmarks are facts that must be true at some point during the execution of any solution plan.

**Definition 1** *Given a planning task $(A, I, G)$. A fact $L$ is a* landmark *if for all $P = \langle a_1, \ldots, a_n \rangle \in A^*, G \subseteq Result(I, P) : L \in Result(I, \langle a_1, \ldots, a_i \rangle)$ for some $0 \leq i \leq n$.*

Note that in an unsolvable task *all* facts are landmarks (by universal quantification over the empty set of solution plans in the above definition). The definition thus only makes sense if the task at hand is solvable. Indeed, while our landmark techniques can help a planning algorithm to find a solution plan faster (as we will see later), they are not useful for proving unsolvability. The reasonable orders we will introduce are based on heuristic notions that make sense intuitively, but that are not *mandatory* in the sense that every solution plan obeys them, or even in the sense that there exists a solution plan that obeys them. Details on this topic are given with the individual concepts below. We remark that we make these observations only to clarify the meaning of our definitions. Given the way we use the landmarks information for planning, for our purposes it is not essential if or if not an ordering constraint is mandatory. Our search control loop only suggests to the planner what might be good to achieve next, it does not force the planner to do so (see Section 5).

Initial and goal facts are trivially landmarks: set $i$ to 0 respectively $n$ in Definition 1. In general, it is PSPACE-complete to decide whether a fact is a landmark or not.

**Theorem 1** *Let* LANDMARK *denote the following problem: given a planning task $(A, I, G)$, and a fact $L$; is $L$ a landmark?*

*Deciding* LANDMARK *is PSPACE-complete.*

**Proof Sketch:** PSPACE-hardness follows by a straightforward reduction of the complement of PLANSAT– the decision problem of whether there exists a solution plan to a given arbitrary STRIPS task (Bylander, 1994) – to the problem of deciding LANDMARK. PSPACE-membership follows vice versa. □

Full proofs are in Appendix A. One of the most elementary ordering relations between a pair $L$ and $L'$ of landmarks is the following. In any action sequence that makes $L'$ true in some state, $L$ is true in the immediate preceding state. Typically, a fact $L$ *is* a landmark because it is ordered in this way before some other landmark $L'$. The reason is typically that $L$ is a necessary prerequisite – a shared precondition – for achieving $L'$. We will exploit this for our approximation techniques in Section 4.

**Definition 2** *Given a planning task* $(A, I, G)$*, and two facts* $L$ *and* $L'$*. There is a* necessary order *between* $L$ *and* $L'$*, written* $L \rightarrow_n L'$*, if* $L' \notin I$*, and for all* $P = \langle a_1, \dots, a_n \rangle \in A^*$*: if* $L' \in Result(I, \langle a_1, \dots, a_n \rangle)$ *then* $L \in Result(I, \langle a_1, \dots, a_{n-1} \rangle)$*.*

The definition allows for arbitrary facts, but the case that we will be interested in is the case where $L$ and $L'$ are landmarks. Note that if $L' \in Result(I, \langle a_1, \dots, a_n \rangle)$ then $n \geq 1$ as $L' \notin I$. The intention behind a necessary order $L \rightarrow_n L'$ is that one must have $L$ true before one can have $L'$ true. So it does not make sense to allow such orders for initial facts $L'$. It is important that $L$ is postulated to be true *directly* before $L'$ – this way, if two facts $L$ and $L''$ are necessarily ordered before the same fact $L'$, one can conclude that $L$ and $L''$ must be true *together* at some point. We make use of this observation in our approximation of reasonable orders (see Section 4.2).

We denote necessary orders, and all the other ordering relations we will introduce, as directed graph edges "$\rightarrow$" rather than with the more usual "$<$" symbol. We do this to avoid confusion about the meaning of our relations. As said earlier, none of the ordering relations we introduce is transitive. (Note that $\rightarrow_n$ would be transitive if $L$ was only postulated to hold sometime before $L'$, not directly before it.)

Necessary orders are mandatory. We say that an action sequence $\langle a_1, \dots, a_n \rangle$ *obeys* an order $L \rightarrow L'$ if the sequence makes $L$ true the first time before it makes $L'$ true the first time. Precisely, $\langle a_1, \dots, a_n \rangle$ obeys $L \rightarrow L'$ if either $L \in I$, or $min\{i \mid L \in add(a_i)\} < min\{i \mid L' \in add(a_i)\}$ where the minimum over an empty set is $\infty$. That is, either $L$ is true initially, or $L'$ is not added at all, or $L$ is added before $L'$. By definition, any action sequence obeys necessary orders. So one does not lose solutions if one forces a planner to obey necessary orders, i.e. if one disallows plans violating the orders. (We reiterate that this is a purely theoretical observation; as said above, our search control does not enforce the found ordering constraints.)

**Theorem 2** *Let* NECESSARY-ORD *denote the following problem: given a planning task* $(A, I, G)$*, and two facts* $L$ *and* $L'$*; does* $L \rightarrow_n L'$ *hold?*

*Deciding* NECESSARY-ORD *is PSPACE-complete.*

**Proof Sketch:** PSPACE-hardness follows by reducing the complement of PLANSAT to NECESSARY-ORD. PSPACE-membership follows with a non-deterministic algorithm that guesses action sequences and checks if there is a counter example to the ordering. □

Another interesting relation are *greedy necessary orders*, a slightly weaker version of the necessary orders above. We postulate not that $L$ is true prior to $L'$ in all action sequences, but only in those action sequences where $L'$ is achieved *for the first time*. These are the orders that we actually approximate and use in our implementation (see Section 4).

**Definition 3** *Given a planning task $(A, I, G)$, and two facts $L$ and $L'$. There is a* greedy necessary order *between $L$ and $L'$, written $L \rightarrow_{gn} L'$, if $L' \notin I$, and for all $P = \langle a_1, \ldots, a_n \rangle \in A^*$: if $L' \in Result(I, \langle a_1, \ldots, a_n \rangle)$ and $L' \notin Result(I, \langle a_1, \ldots, a_i \rangle)$ for $0 \leq i < n$, then $L \in Result(I, \langle a_1, \ldots, a_{n-1} \rangle)$.*

Like above with the necessary orders, the action sequence achieving $L'$ must contain at least one step as $L' \notin I$. Obviously, $\rightarrow_n$ is stronger than $\rightarrow_{gn}$, that is, with $L \rightarrow_n L'$ for two facts $L$ and $L'$, $L \rightarrow_{gn} L'$ follows. Greedy necessary orders are still mandatory in the sense that every action sequence obeys them.

The definition of greedy necessary orders captures the fact that, really, what we are interested in is what happens when we directly achieve $L'$ from the initial state, rather than in some remote part of the state space. The consideration of these more remote parts of the state space, which is inherent in the definition of the non-greedy necessary orders, can make us lose useful information. Consider the *Blocksworld* example in Figure 1. There is a greedy necessary order between *clear(D)* and *clear(C)*, *clear(D)* $\rightarrow_{gn}$ *clear(C)*, but not a necessary order, *clear(D)* $\not\rightarrow_n$ *clear(C)*. If we make *clear(C)* true the first time in an action sequence from the initial state, then the action achieving *clear(C)* will always be *unstack(D C)*, which requires *clear(D)* to be true. On the other hand, there can of course be action sequences which achieve *clear(C)* by different actions (*unstack(A C)*, for example). But reaching a state where *clear(C)* can be achieved by such an action involves unstacking D from C, and thus achieving *clear(C)*, in the first place. We will see later (Section 4.2) that the order *clear(D)* $\rightarrow_{gn}$ *clear(C)* can be used to make the important inference that *clear(C)* is reasonably ordered before *on(B D)*.

More generally, the definition of greedy necessary orders is made from the perspective that we are interested in ordering the *first occurence* of the facts $L$ in our desired solution plan. All definitions and algorithms in the rest of this paper are designed from this same perspective. Since a fact might (have to) be made true several times in a solution plan, one could just as well focus on ordering the fact's last occurence, or any occurence, or several occurences of it. We chose to focus on the first occurences of facts mainly in order to keep things simple. It seems very hard to say anything useful a priori about exactly how often and when some fact will need to become true in a plan. The "greedy assumption" that our approach thus makes is that all the landmarks need to be achieved only once, and that it is best to achieve them as early as possible. Of course this assumption is not always justified, and may lead to difficulties, such as e.g. cycles in the generated LGG (see also Sections 4.4 and 6.8). Generalising our approach to take account of several occurences of the same fact is an open research topic.

**Theorem 3** *Let* GREEDY-NECESSARY-ORD *denote the following problem: given a planning task $(A, I, G)$, and two facts $L$ and $L'$; does $L \rightarrow_{gn} L'$ hold?*

*Deciding* GREEDY-NECESSARY-ORD *is PSPACE-complete.*

**Proof Sketch:** By a minor modification of the proof to Theorem 2. □

### 3.2 Reasonable Orders

Reasonable orders were first introduced by Koehler and Hoffmann (2000), for top level goals. We extend their definition, in a slightly revised way, to landmarks.

Let us first reiterate what the idea of reasonable orders was originally. The idea introduced by Koehler and Hoffmann is this. If the planner is in a state $s$ where one goal $L'$ has just been achieved, but another goal $L$ is still false, and $L'$ must be destroyed in order to achieve $L$, then it might have been better to achieve $L$ first: to get to a goal state from $s$, the planner will have to delete and re-achieve $L'$. If the same situation arises in *all* states $s$ where $L'$ has just been achieved but $L$ is false, then it seems reasonable to generally introduce an ordering constraint $L \rightarrow L'$, indicating that $L$ should be achieved prior to $L'$.

The classical example for two facts with a reasonable ordering constraint are *on* relations in *Blocksworld*, where *on(B, C)* is reasonably ordered before *on(A, B)* whenever the goal is to have both facts true in the goal state. Obviously, if one achieves *on(A, B)* first then one has to unstack A again in order to achieve *on(B, C)*.

Think about an unmodified application of Koehler and Hoffmann's definition to the case of landmarks. Consider a state $s$ where we have a landmark $L'$, but not another landmark $L$, and achieving $L$ involves deleting $L'$. Does it matter? It might be that we do not need to achieve $L$ from $s$ anyway. It might also be that we do not need $L'$ anymore once we have achieved $L$. In both cases, there is no need to delete and re-achieve $L'$, and it does *not* appear reasonable to introduce the constraint $L \rightarrow L'$. The question is, under which circumstances is it reasonable? The answer is given by the two mentioned counter-examples. The situation matters if 1. we need to achieve $L$ from $s$, and 2. we must re-achieve $L'$ again afterwards. Both conditions are trivially fulfilled when $L$ and $L'$ are top level goals. Our definition below makes sure they hold for the landmarks $L$ and $L'$ in question.

We say that there is a reasonable ordering constraint between two landmarks $L$ and $L'$ if, starting from any state where $L'$ was achieved before $L$: $L'$ must be true at some point later than the achievement of $L$; and one must delete $L'$ on the way to $L$. Formally, we first define the "set of states where $L'$ was achieved before $L$", then we define what it means that "$L'$ must be true at some point later than the achievement of $L$", then based on that we define what reasonable orders are.

**Definition 4** *Given a planning task $(A, I, G)$, and two facts $L$ and $L'$.*

1. *By $S_{(L',\neg L)}$, we denote the set of states $s$ such that there exists $P = \langle a_1, \ldots, a_n \rangle \in A^*$, $s = Result(I, P)$, $L' \in add(a_n)$, and $L \notin Result(I, \langle a_1, \ldots, a_i \rangle)$ for $0 \leq i \leq n$.*

2. *$L'$ is in the* aftermath *of $L$ if, for all states $s \in S_{(L',\neg L)}$, and all solution plans $P = \langle a_1, \ldots, a_n \rangle \in A^*$ from $s$, $G \subseteq Result(s, P)$, there are $1 \leq i \leq j \leq n$ such that $L \in Result(s, \langle a_1, \ldots, a_i \rangle)$ and $L' \in Result(s, \langle a_1, \ldots, a_j \rangle)$.*

3. *There is a* reasonable order *between $L$ and $L'$, written $L \rightarrow_r L'$, if $L'$ is in the aftermath of $L$, and*

$$\forall\, s \in S_{(L',\neg L)} : \forall\, P \in A^* : \; L \in Result(s, P) \Rightarrow \exists\, a \in P : L' \in del(a)$$

Let us explain this definition, and how it differs from Koehler and Hoffmann's original one.

1. $S_{(L',\neg L)}$ contains the states where $L'$ was just added, but $L$ was not true yet. These are the states we consider: we are interested to know if, from every state $s \in S_{(L',\neg L)}$, we will have to delete and re-achieve $L'$. In Koehler and Hoffmann's original definition, $S_{(L',\neg L)}$ contained more states, namely all those states $s$ where $L'$ was just added but $L \notin s$. This definition allowed cases where $L$ was achieved already but was deleted again. Our revised definition captures better the intuition that we want to consider all states where $L'$ was achieved *before* $L$. The revised definition also makes sure that, for a landmark $L$, any solution plan starting from $s \in S_{(L',\neg L)}$ must achieve $L$ at some point.

2. The definition of the aftermath relation just says that, in a solution plan starting from $s \in S_{(L',\neg L)}$, $L'$ must be true simultaneously with $L$, or at some later time point. Koehler and Hoffmann didn't need such a definition since this condition is trivially fulfilled for top level goals.

3. The definition of $L \rightarrow_r L'$ then says that, from every $s \in S_{(L',\neg L)}$, every action sequence achieving $L$ deletes $L'$ at some point. With the additional postulation that $L'$ is in the aftermath of $L$, this implies that from every $s \in S_{(L',\neg L)}$ one needs to delete and re-achieve $L'$. Koehler and Hoffmann's definition here is identical except that they do not need to postulate the aftermath relation.

Because in their definition $S_{(L',\neg L)}$ contains more states, and top level goals are trivially in the aftermath of each other, Koehler and Hoffmann's $\rightarrow_r$ definition is stronger than ours, i.e. $L \rightarrow_r L'$ in the Koehler and Hoffmann sense implies $L \rightarrow_r L'$ as defined above (we give an example below where our, but not the Koehler and Hoffmann $L \rightarrow_r L'$ relation holds).[4]

It is important to note that reasonable orders are *not* mandatory. An order $L \rightarrow_r L'$ only says that, if we achieve $L'$ before $L$, we will need to delete and re-achieve $L'$. This *might* mean that achieving $L'$ before $L$ is wasted effort. But there are cases where, in the process of achieving some landmark $L$, one has no other choice but to achieve, delete, and re-achieve a landmark $L'$. In the *Towers of Hanoi* domain, for example, this is the case for nearly all pairs of top level goals – namely, for all those pairs of goals that say that ($L'$) disc $i$ must be located on disc $i+1$, and ($L$) disc $i+1$ must be located on disc $i+2$. In such a situation, forcing a planner to obey the order $L \rightarrow_r L'$ cuts out all solution paths. One can also easily construct cases where $L \rightarrow_r L'$ *and* $L' \rightarrow_r L$ hold for goals $L$ and $L'$ (that can not be achieved simultaneously). Consider the following example. There are the

4. Note that an order $L \rightarrow_r L'$ intends to tell us that we should not achieve $L'$ *before* $L$. This leaves open the option to achieve $L$ and $L'$ simultaneously. In that sense, our definition (given above in Section 3.1) of what it means to obey an order $L \rightarrow L'$, namely to add $L$ strictly before $L'$, is a bit too restrictive. In our experience, the restriction is irrelevant in practice. In none of the many benchmarks we tried did we observe facts that were reasonably ordered (ordered at all, in fact) relative to each other and that could be achieved with the same action – remember that we consider the sequential planning setting. We remark that one can easily adapt our framework to take account of simultaneous achievement of $L$ and $L'$. No changes are needed except in the approximation of obedient reasonable orders, which would become slightly more complicated, see Section 4.3.

seven facts $L$, $L'$, $P_1$, $P_2$, $P_2'$, $P_3$, and $P_3'$. Initially only $P_1$ is true, and the goal is to have $L$ and $L'$. The actions are:

$$\begin{array}{lclll} \text{name} & & (pre, & add, & del) \\ \mathbf{op}L_1 & = & (\{P_1\}, & \{L, P_2\}, & \{P_1\}) \\ \mathbf{op}L_1' & = & (\{P_1\}, & \{L', P_2'\}, & \{P_1\}) \\ \mathbf{op}L_2 & = & (\{P_2'\}, & \{L, P_3\}, & \{L', P_2'\}) \\ \mathbf{op}L_2' & = & (\{P_2\}, & \{L', P_3'\}, & \{L, P_2\}) \\ \mathbf{op}L_3 & = & (\{P_3'\}, & \{L\}, & \{P_3'\}) \\ \mathbf{op}L_3' & = & (\{P_3\}, & \{L'\}, & \{P_3\}) \end{array}$$

Figure 2 shows the state space of the example. There are exactly two solution paths, $\langle$ $\mathbf{op}L_1$, $\mathbf{op}L_2'$, $\mathbf{op}L_3$ $\rangle$ and $\langle$ $\mathbf{op}L_1'$, $\mathbf{op}L_2$, $\mathbf{op}L_3'$ $\rangle$. The first of these paths achieves, deletes, and re-achieves $L$, the second one does the same with $L'$. $S_{(L',\neg L)}$ contains the single state that results from applying $\mathbf{op}L_1'$ to the initial state. From that state, one has to apply $\mathbf{op}L_2$ in order to achieve $L$, deleting $L'$, so $L \rightarrow_r L'$ holds. Similarly, it can be seen that $L' \rightarrow_r L$ holds. Note that either solution path disobeys one of the two reasonable orders.

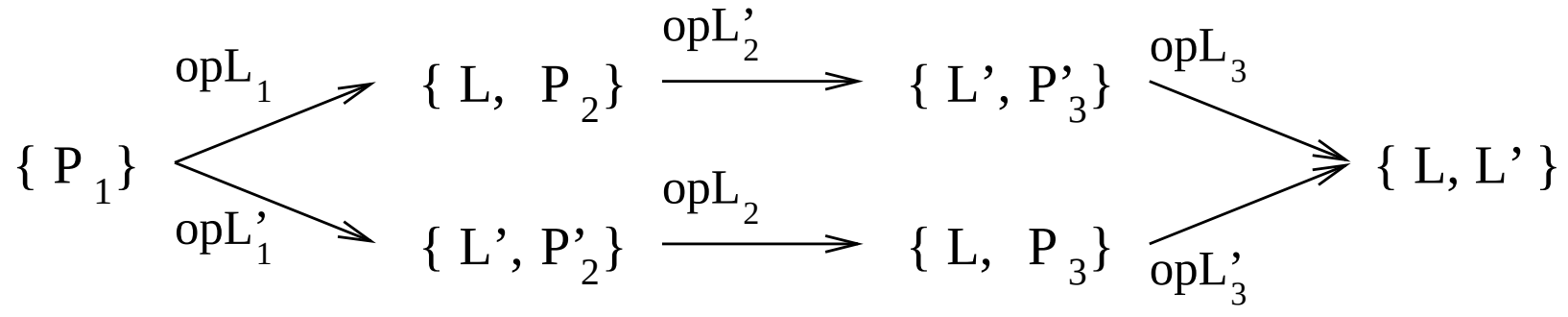

Figure 2: State space of the example.

We reiterate that the above are purely theoretical observations made to clarify the meaning of our definitions. Our search control does not enforce the found ordering constraints, it only suggests them to the planner.

While reasonable orders $L \rightarrow_r L'$ are not mandatory, they can help to reduce search effort in those cases where achieving $L'$ before $L$ *does* imply wasted effort. Our *Blocksworld* example from Figure 1 constitutes such a case. In the example, it makes no sense to stack $B$ onto $D$ while $D$ is still located on $C$, because $C$ has to end up on top of $A$. By Definition 4, *clear(C)* $\rightarrow_r$ *on(B D)* holds: $S_{(on(BD),\neg clear(C))}$ contains only states where $B$ has been stacked onto $D$, but $D$ is still on top of $C$. From these states, one must delete *on(B D)* in order to achieve *clear(C)*. Further, *on(B D)* is a top-level goal so it is in the aftermath of *clear(C)*, and *clear(C)* $\rightarrow_r$ *on(B D)* follows. The order does not hold in terms of Koehler and Hoffmann's definition, because there the $S_{(on(BD),\neg clear(C))}$ state set also contains states where $D$ was already removed from $C$.

Like the previous decision problems, those related to the aftermath relation and to reasonable orders are PSPACE-complete.

**Theorem 4** *Let* AFTERMATH *denote the following problem: given a planning task* $(A, I, G)$, *and two facts L and* $L'$; *is* $L'$ *in the aftermath of L?*

*Deciding* AFTERMATH *is PSPACE-complete.*

**Proof Sketch:** PSPACE-hardness follows by reducing the complement of PLANSAT to AFTERMATH. PSPACE-membership follows by a non-deterministic algorithm that guesses counter examples. □

**Theorem 5** *Let* REASONABLE-ORD *denote the following problem: given a planning task* $(A, I, G)$*, and two facts* $L$ *and* $L'$ *such that* $L'$ *is in the aftermath of* $L$*; does* $L \rightarrow_r L'$ *hold?*

*Deciding* REASONABLE-ORD *is PSPACE-complete.*

**Proof Sketch:** PSPACE-hardness follows by reducing the complement of PLANSAT to REASONABLE-ORD, with the same construction as used by Koehler and Hoffmann (2000) for the original definition of reasonable orders. PSPACE-membership follows by a non-deterministic algorithm that guesses counter examples. □

### 3.3 Obedient Reasonable Orders

Say we already have a set $O$ of reasonable ordering constraints $L \rightarrow_r L'$. The question we focus on in the section at hand is, if a planner commits to obey all the constraints in $O$, do other reasonable orders arise? The answer is, yes, there might.

Consider the following situation. Say we got landmarks $L$ and $L'$, such that we must delete $L'$ in order to achieve $L$. Also, there is a third landmark $L''$ such that $L' \rightarrow_n L''$ and $L \rightarrow_r L''$. Now, if the order $L \rightarrow L''$ was necessary, $L \rightarrow_n L''$, then we would have a reasonable order $L \rightarrow_r L'$: $L$ and $L'$ would need to be true together immediately prior to the achievement of $L''$, so $L'$ would be in the aftermath of $L$. However, the ordering constraint $L \rightarrow L''$ is "only" reasonable so there is no guarantee that a solution plan will obey it. A plan can choose to achieve $L'$ before $L''$ before $L$, and thereby avoid deletion and re-achievement of $L'$. But if we enforce the ordering constraint $L \rightarrow_r L''$, disallowing plans that do not obey it, then achieving $L'$ before $L$ leads to deletion and re-achievement of $L'$ and is thus not reasonable.

With the above, the idea we pursue now is to define a weaker form of reasonable orders, which are obedient in the sense that they only arise if one commits to a given set $O$ of (previously computed) reasonable ordering constraints. In our experiments, using (an approximation of) such obedient reasonable orders, on top of the reasonable orders themselves, resulted in significantly better planner performance in a few domains (such as the Blocksworld), and made no difference in the other domains. Summarised, what we do is, we start from the set $O$ of reasonable orders already computed by our approximations, and then insert new orders that are reasonable given one commits to obey the constraints in $O$. We do this just once, i.e. we do not compute a fixpoint. The details are in Section 4.3. Right now, we define what obedient reasonable orders are.

The definition of obedient reasonable orders is almost the same as that of reasonable orders. The only difference lies in that we consider only action sequences that are *obedient* in the sense that they obey all ordering constraints in the given set $O$. The definition of when an action sequence $\langle a_1, \ldots, a_n \rangle$ obeys an order $L \rightarrow L'$ was already given above: if either $L \in I$, or $min\{i \mid L \in add(a_i)\} < min\{i \mid L' \in add(a_i)\}$ where the minimum over an empty set is $\infty$.

**Definition 5** *Given a planning task $(A, I, G)$, a set $O$ of reasonable ordering constraints, and two facts $L$ and $L'$.*

1. *By $S^O_{(L',\neg L)}$, we denote the set of states $s$ such that there exists an obedient action sequence $P = \langle a_1, \ldots, a_n \rangle \in A^*$, with $s = Result(I, P)$, $L' \in add(a_n)$, and $L \notin Result(I, \langle a_1, \ldots, a_i \rangle)$ for $0 \leq i \leq n$.*

2. *$L'$ is in the* obedient aftermath *of $L$ if, for all states $s \in S^O_{(L',\neg L)}$, and all obedient solution plans $P = \langle a_1, \ldots, a_n \rangle \in A^*$, $G \subseteq Result(I, P)$, where $s = Result(I, \langle a_1, \ldots, a_k \rangle)$, there are $k \leq i \leq j \leq n$ such that $L \in Result(I, \langle a_1, \ldots, a_i \rangle)$ and $L' \in Result(I, \langle a_1, \ldots, a_j \rangle)$.*

3. *There is an* obedient reasonable order *between $L$ and $L'$, written $L \rightarrow^O_r L'$, if and only if $L'$ is in the obedient aftermath of $L$, and*

$$\forall\, s \in S^O_{(L',\neg L)} : \forall\, P \in A^* :\; L \in Result(s, P) \Rightarrow \exists\, a \in P : L' \in del(a)$$

This definition is very similar to Definition 4 and thus should be self-explanatory, in its formal aspects. The definition of the aftermath relation looks a little more complicated because the solution plan $P$ starts from the initial state, not from $s$ as in Definition 4, and reaches $s$ with action $a_k$. This is just a minor technical device to cover the case where, for some of the $L_1 \rightarrow_r L_2$ constraints in $O$, $L_1$ is contained in $s$ already (and thus does not need to be added after $s$ in order to obey $L_1 \rightarrow_r L_2$). Note that, in part 3 of the definition, the action sequences $P$ achieving $L$ are not required to be obedient. While it would make sense to impose this requirement, our approximation techniques (that will be introduced in Section 4.3) only take account of $O$ in the computation of the aftermath relation anyway. It is an open question how our other approximation techniques could be made to take account of $O$.

We remark that the modified definitions do not change the computational complexity of the corresponding decision problems.[5] As a quick illustration of the new definitions, reconsider the situation described above. There, $L'$ is not in the aftermath of $L$, but in the obedient aftermath of $L$ because all action sequences that obey the constraint $L \rightarrow_r L''$ make $L'$ true at a point simultaneously with or behind $L$ (namely immediately prior to $L''$, assuming that there is no action that adds both $L$ and $L''$). As $L'$ must be deleted in order to achieve $L$, we obtain the ordering $L \rightarrow^{\{L \rightarrow_r L''\}}_r L'$. That is, if the planner obeys the constraint $L \rightarrow_r L''$ then it is reasonable to also order $L$ before $L'$.

Just like the reasonable orders, the obedient reasonable orders are not mandatory. Enforcing an obedient reasonable order can cut out all solution paths. The reason is the same as for the reasonable orders. An order $L \rightarrow^O_r L'$ only says that, given we want to obey $O$, achieving $L'$ before $L$ implies deletion and re-achievement of $L'$. If this really means that achieving $L'$ before $L$ is wasted effort, the order tells us nothing about. Consider the following example. There are the ten facts $L$, $L'$, $L''$, $P$, $A_1$, $A_2$, $A_3$, $B_1$, $B_2$, and $B_3$.

5. For the obedient aftermath relation, minor modifications of the proof to Theorem 4 suffice. PSPACE-hardness follows by using the empty set of ordering constraints. PSPACE-membership follows by extending the non-deterministic decision algorithm with flags that check if the ordering constraints are adhered to. Similarly for obedient reasonable orders and the proof to Theorem 5.

Initially only $P$ is true, and the goal is to have $L$, $L'$, and $L''$. The construction is made so that $L \rightarrow_r L''$, and $L \not\rightarrow_r L'$ but $L \rightarrow_r^{\{L \rightarrow_r L''\}} L'$. Enforcing $L \rightarrow_r^{\{L \rightarrow_r L''\}} L'$ renders the task unsolvable. The actions are:

| name | | (*pre*, | *add*, | *del*) |
|---|---|---|---|---|
| $\mathbf{op}A$ | = | $(\{P\}$, | $\{A_1\}$, | $\{P\})$ |
| $\mathbf{op}B$ | = | $(\{P\}$, | $\{B_1\}$, | $\{P\})$ |
| $\mathbf{op}A_1$ | = | $(\{A_1\}$, | $\{L', L'', A_2\}$, | $\{A_1\})$ |
| $\mathbf{op}A_2$ | = | $(\{A_2\}$, | $\{L, A_3\}$, | $\{L'', A_2\})$ |
| $\mathbf{op}A_3$ | = | $(\{A_3\}$, | $\{L''\}$, | $\{A_3\})$ |
| $\mathbf{op}B_1$ | = | $(\{B_1\}$, | $\{L', B_2\}$, | $\{B_1\})$ |
| $\mathbf{op}B_2$ | = | $(\{B_2\}$, | $\{L, B_3\}$, | $\{L', B_2\})$ |
| $\mathbf{op}B_3$ | = | $(\{B_3\}$, | $\{L', L''\}$, | $\{B_3\})$ |

Figure 3 shows the state space of the example. One has to choose one out of two options. First, one applies $\mathbf{op}A$ to the initial state and then proceeds with $\mathbf{op}A_1$, $\mathbf{op}A_2$, and $\mathbf{op}A_3$. Second, one applies $\mathbf{op}B$ to the initial state and proceeds with $\mathbf{op}B_1$, $\mathbf{op}B_2$, and $\mathbf{op}B_3$. The first option is the only one where $L''$ becomes true before $L$. One has to delete $L''$ with $\mathbf{op}A_2$, and re-achieve it with $\mathbf{op}A_3$. For this reason, the order $L \rightarrow_r L''$ holds. The order $L \rightarrow_r L'$ does not hold because if one chooses the first option then $L'$ becomes true prior to $L$, and is never deleted. However, committing to the order $L \rightarrow_r L''$ means excluding the first option. In the second option, $L'$ becomes true before $L$, and must then be deleted and re-achieved, so we get the order $L \rightarrow_r^{\{L \rightarrow_r L''\}} L'$. But there is no solution plan that obeys this order because there is no way to make $L$ true before (or, even, simultaneously with) $L'$.

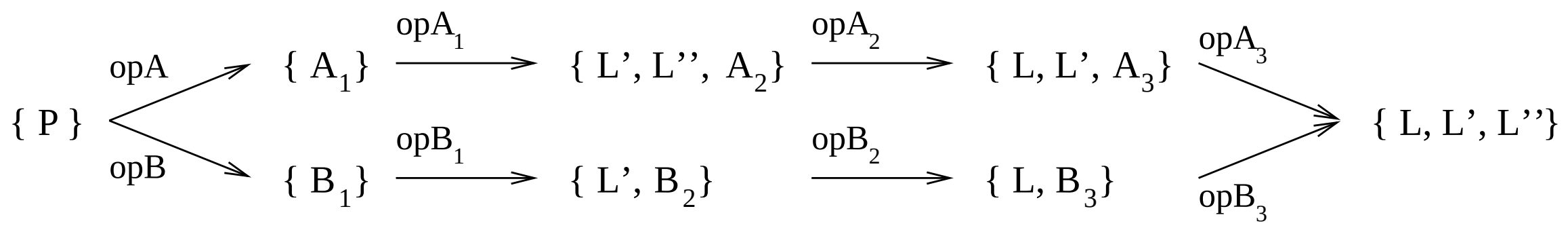

Figure 3: State space of the example.

## 4. How to Find Ordered Landmarks

We now describe our methods to find landmarks in a given planning task, and to approximate their inherent ordering constraints. The result of the process is a directed graph in the obvious way, the landmarks generation graph (LGG). Section 4.1 describes how we find landmarks, and how we approximate greedy necessary orders between them. Section 4.2 gives a sufficient criterion for reasonable orders, based on greedy necessary orders and fact inconsistencies, and describes how we use the criterion for approximating reasonable orders. Section 4.3 adapts this technology to obedient reasonable orders. Section 4.4 describes our handling of cycles in the LGG, and Section 4.5 describes a preliminary form of "lookahead" orders that we have also implemented and used.

### 4.1 Finding Landmarks and Approximating Greedy Necessary Orders

We find (a subset of the) landmarks in a planning task, and approximate the greedy necessary orders between them, both in one process. The process is split into two parts:

1. **Compute an LGG of landmark candidates together with approximated greedy necessary orders between them.** This is done with a backchaining process. The goals form the first landmark candidates. Then, for any candidate $L'$, the "earliest" actions that can be used to achieve $L'$ are considered. Here, "early" is a greedy approximation of reachability from the initial state. The actions are analysed to see if they have shared precondition facts $L$ – facts that must be true before executing any of the actions. These facts $L$ become new candidates if they have not already been processed, and the orders $L \rightarrow_{gn} L'$ are introduced. The process is iterated until there are no new candidates. (Due to the greedy selection of actions, $L$/the order $L \rightarrow_{gn} L'$ is not *proved* to be a landmark/a greedy necessary order.)

2. **Remove from the LGG the candidates (and their incident edges) that can not be proved to be landmarks.** This is done by evaluating a sufficient condition on each candidate $L$ in the LGG. The condition ignores all actions that add $L$, and asks if a relaxed version of the task is still solvable. If not, $L$ is proved to be a landmark. (Any relaxation can be used in principle; we use the relaxation that ignores delete lists as in McDermott, 1999 and Bonet & Geffner, 2001.)

The next two subsections focus on these two steps in turn.

#### 4.1.1 Landmark Candidates

We give pseudo-code for our approximation algorithm below. As said, we make the algorithm greedy by using an approximation of reachability from the initial state. The approximation we use is a *relaxed planning graph* (Hoffmann & Nebel, 2001), short RPG. Let us explain this data structure first. An RPG is built just like a planning graph (Blum & Furst, 1997), except that the delete lists of all actions are ignored; as a result, there are no mutex relations in the graph. The RPG thus is a sequence $P_0, A_0, P_1, A_1, \ldots, P_{m-1}, A_{m-1}, P_m$ of proposition sets (layers) $P_i$ and action sets (layers) $A_i$. $P_0$ contains the facts that are true in the initial state, $A_0$ contains those actions whose preconditions are reached (contained) in $P_0$, $P_1$ contains $P_0$ plus the add effects of the actions in $A_0$, and so on. We have $P_i \subseteq P_{i+1}$ and $A_i \subseteq A_{i+1}$ for all $i$. If the relaxed task (without delete lists) is unsolvable, then the RPG reaches a fixpoint before reaching the goal facts, thereby proving unsolvability. If the relaxed task is solvable, then eventually a layer $P_m$ containing the goal facts will be reached.[6]

6. Note that the RPG thus decides solvability of the relaxed planning task. Indeed, building an RPG is a variation of the algorithm given by Bylander (1994) to prove that plan existence is polynomial in the absence of delete lists.

An RPG encodes an over-approximation of reachability in the planning task. We define the *level* of a fact/action to be the index of the first proposition/action layer that contains the fact/action. Then, if the level of a fact/action is $l$, one must apply at least $l$ parallel action steps from the initial state before the fact becomes true/the action becomes applicable. (The fact/action level corresponds to the "$h^1$" heuristic defined by Haslum & Geffner, 2000.) We use this over-approximation of reachability to insert some greediness into our approximation of "greedy" necessary orders (more below). The approximation process proceeds as shown in Figure 4.

initialise the LGG to $(G, \emptyset)$, and set $C := G$
**while** $C \neq \emptyset$ **do**
  set $C' := \emptyset$
  **for** all $L' \in C, level(L') \neq 0$ **do**
    let $A$ be the set of all actions $a$ such that $L' \in add(a)$, and $level(a) = level(L') - 1$  $(*)$
    **for** all facts $L$ such that $\forall a \in A : L \in pre(a)$ **do**
      if $L$ is not yet a node in the LGG, set $C' := C' \cup \{L\}$
      if $L$ is not yet a node in the LGG, then insert that node
      if $L \rightarrow_{gn} L'$ is not yet an edge in the LGG, then insert that edge
    **endfor**
  **endfor**
  set $C := C'$
**endwhile**

Figure 4: Landmark candidate generation.

The set of landmark candidates is initialised to comprise the goal facts. Each iteration of the **while**-loop processes all "open" candidates $L'$ – those $L'$ in $C$. Candidates $L'$ with level 0, i.e., initial facts, are not used to produce greedy necessary orders and new landmark candidates, because after all such $L'$ are already true. For the other open candidates $L'$, the set $A$ comprises all those actions at the level below $L'$ that can be used to achieve $L'$. Note that these are the earliest possible achievers of $L'$ in the RPG, or else the level of $L'$ would be lower. We take as the new landmark candidates those facts $L$ that every action in $A$ requires as a precondition, and update the LGG and the set of open candidates accordingly. Independently of the $(*)$ step, the algorithm terminates because there are only finitely many facts. Because we use the RPG level test in step $(*)$, the levels of the new candidates $L$ are strictly lower than the level of $L'$, and the **while**-loop terminates after at most $m$ iterations where $m$ is the index of the topmost proposition layer in the RPG.

If we skipped the test for the RPG level at the point in the algorithm marked $(*)$, then the new candidates $L$ would be proved landmarks, and the generated orders would be proved to be necessary and thus also greedy necessary. Obviously, if *all* actions that can achieve a landmark $L'$ require $L$ to be true, then $L$ is a landmark that must be true immediately prior to achieving $L'$. Restricting the choice of $L'$ achievers with the RPG level test, the found landmarks and orders may be unsound. Consider the following example, where we want to move from city $A$ to city $D$ on the road map shown in Figure 5, using a standard *move* operator.

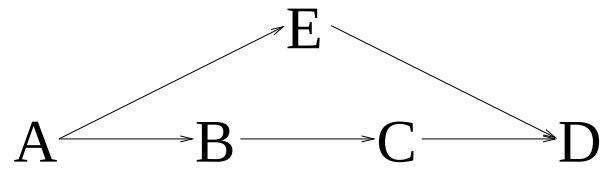

Figure 5: An example road map.

The above algorithm will come up with the following LGG: $\{at(A), at(E), at(D)\}, \{at(A) \rightarrow_{gn} at(E), at(E) \rightarrow_{gn} at(D)\}$ – the RPG is only built until the goals are reached the first time, which happens in this example before *move(C D)* comes in. However, the action sequence $\langle$*move(A B), move(B C), move(C D)*$\rangle$ achieves *at(D)* without making *at(E)* true. Therefore, *at(E)* is not really a landmark, and $at(E) \rightarrow_{gn} at(D)$ is not really a greedy necessary order.

By restricting our choice of $L'$ achievers with the RPG level test at step $(*)$ in Figure 4, as said we intend to insert greediness into our approximation of greedy necessary orders. The generated orders $L \rightarrow_{gn} L'$ are only guaranteed to be sound if, in the RPG, the set of earliest achievers of $L'$ contains all actions that can be used to make $L'$ true for the first time from the initial state. Of course, it is hard to exactly compute that latter set of actions, and also it is highly non-trivial – if possible at all – to find general conditions on when the earliest achievers in the RPG contain all these actions. In the road map example above, the actions that can achieve *at(D)* for the first time are *move(C D)* and *move(E D)*, but the only earliest achiever in the RPG is *move(E D)*. This leads to the unsound $at(E) \rightarrow_{gn} at(D)$ order. In the following example taken from the well-known *Logistics* domain, the earliest achievers of $L'$ *do* contain all actions that can make $L'$ true for the first time. Say $L' = at(P\ A)$ requires package $P$ to be at the airport $A$ of its origin city, and $P$ is not at this airport initially. The actions that can achieve $L'$ are to unload $P$ from the local truck $T$, or to unload it from any airplane. The only earliest achiever in the RPG is the unload from $T$, and indeed that's the only action that can achieve $L'$ for the first time – in order to get the package into an airplane, the package has to arrive at the airport in the first place. Our approximation process correctly generates the new landmark candidate *in(P T)* as well as the greedy necessary order $in(P\ T) \rightarrow_{gn} at(P\ A)$. Note that $in(P\ T) \not\rightarrow_{n} at(P\ A)$.

We show below in Section 4.1.2 how we re-establish the soundness of the landmark candidates, removing candidates (and their associated orders) that are not provably landmarks. We did not find a way to provably re-establish the soundness of the generated greedy necessary orders, and unsound orders may stay in the LGG, potentially also causing the inference of unsound reasonable/obedient reasonable orders (see the sections below). We did observe such unsoundness in a few domains during our experiments (individual discussions are in Section 6). We remark the following.

1. While unsound approximated $L \rightarrow_{gn} L'$ orders are not valid with respect to Definition 3, they still make some sense intuitively. They are generated because $L$ is in the preconditions of all actions that are the first ones in the RPG to achieve $L'$. This means that going to $L'$ via $L$ is probably a good option, in terms of distance from the initial state.

2. Unless $L$ is a landmark for some other reason (than for the unsound order $L \rightarrow_{gn} L'$), landmark verification will remove $L$, and in particular the order $L \rightarrow_{gn} L'$, from the LGG (see the discussion of the Figure 5 example below in the section about landmark verification).

3. As said before, our search control does not enforce the orders in the LGG, it only suggests them to the planner. So even if there is no plan that obeys an order in the LGG, this does not mean that our search control will make the planner fail.

4. If we were to extract only provably necessary orders, by not using the RPG level test, we would miss the information that lies in those $\rightarrow_{gn}$ orders that are not $\rightarrow_n$ orders.

For these reasons, in particular for the last one, we concentrated on the potentially unsound RPG-based approximation in our experiments. We also ran some comparative tests to the "safe" strategy without the RPG level test, in domains where the RPG produced unsound orders. See the details in Section 6.

One case where an $\rightarrow_{gn}$ order, that is not an $\rightarrow_n$ order, contains potentially useful information, is the *Logistics* example given above. Another case is the aforementioned order *clear(D)* $\rightarrow_{gn}$ *clear(C)* in our running *Blocksworld* example from Figure 1. To conclude this subsection, let us have a look at what our approximation algorithm from Figure 4 does in that example. The RPG for the example is summarised in Figure 6.

| $P_0$ | $A_0$ | $P_1$ | $A_1$ | $P_2$ | $A_2$ | $P_3$ |
|---|---|---|---|---|---|---|
| on-table(A) | pick-up(A) | holding(A) | stack(B A) | on(B A) | **stack(C A)** | **on(C A)** |
| **on-table(B)** | **pick-up(B)** | **holding(B)** | **stack(B D)** | **on(B D)** | stack(C B) | on(C B) |
| **on-table(C)** | **unstack(D C)** | holding(D) | stack(B C) | on(B C) | stack(C D) | on(C D) |
| **on(D C)** | | **clear(C)** | put-down(B) | ... | ... | ... |
| **clear(A)** | | | ... | | | |
| **clear(B)** | | | **pick-up(C)** | **holding(C)** | | |
| **clear(D)** | | | ... | ... | | |
| **arm-empty()** | | | | | | |

Figure 6: Summarised RPG for the illustrative *Blocksworld* example from Figure 1.

As we explained above, the extraction process starts by considering the goals *on(C A)* and *on(B D)* as landmark candidates. The RPG level of *on(C A)* is 3, the level of *on(B D)* is 2. There is only one action with level 2 that achieves *on(C A)*: *stack(C A)*. So, *holding(C)* (level 2) and *clear(A)* (level 0) are new candidates. The new LGG is: ({*on(C A)*,*on(B D)*,*holding(C)*,*clear(A)*}, {*holding(C)* $\rightarrow_{gn}$ *on(C A)*, *clear(A)* $\rightarrow_{gn}$ *on(C A)*}). Processing *on(B D)*, we find that its only earliest achiever is *stack(B D)*, and we generate the new candidates *holding(B)* (level 1) and *clear(D)* (level 0) with the respective edges. In the next iteration, *holding(C)* (level 2) produces the new candidates *clear(C)* (level 1), *on-table(C)* (level 0), and *arm-empty()* (level 0) by the achiever *pick-up(C)*; and *holding(B)* (level 1) produces the new candidates *on-table(B)* (level 0) and *clear(B)* (level 0) by the achiever *pick-up(B)*. In the third and final iteration of the algorithm, *clear(C)* (level 1) produces the new candidate *on(D C)* (level 0) by the achiever *unstack(D C)*. The process ends up with the LGG as shown in Figure 7 (the edges in the depicted graph are all directed from bottom to top). Fact sets of which our LGG suggests that they have to be true *together* at some point – because they are either top level goals, or $\rightarrow_{gn}$ ordered before the same

fact – are grouped together in boxes. As said before, this information is important for the approximation of reasonable orders described below.

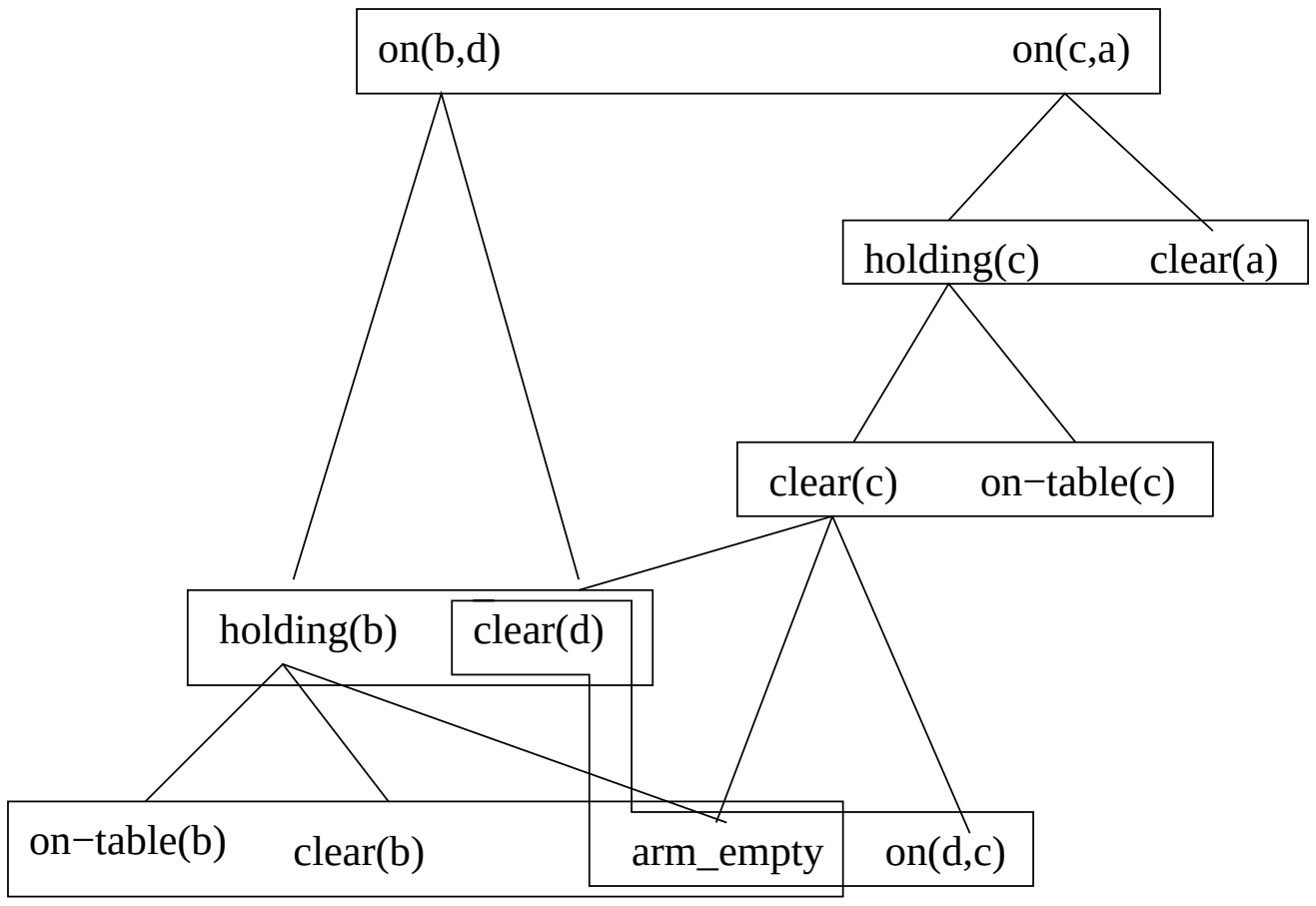

Figure 7: LGG for the illustrative *Blocksworld* task, containing the found landmarks and $\rightarrow_{gn}$ orders.

### 4.1.2 Landmark Verification

As said before, we verify landmark candidates by evaluating a sufficient condition on them, and throwing away those candidates where the condition fails. The condition we use is the following.

**Proposition 1** *Given a planning task* $(A, I, G)$*, and a fact* $L$*. Define a modified action set* $A_L$ *as follows.*

$$A_L := \{(pre(a), add(a), \emptyset) \mid (pre(a), add(a), del(a)) \in A, L \notin add(a)\}$$

*If* $(A_L, I, G)$ *is unsolvable, then* $L$ *is a landmark in* $(A, I, G)$*.*

Note that the inverse direction of the proposition does not hold – that is, if $L$ is a landmark in $(A, I, G)$ then $(A_L, I, G)$ is not necessarily unsolvable – because ignoring the delete lists simplifies the achievement of the goals. As mentioned earlier, deciding about solvability of planning tasks with empty delete lists can be done in polynomial time by building the RPG. The task is unsolvable iff the RPG can't reach the goals. So our landmark verification process looks at all landmark candidates in turn. Candidates that are top level goals or initial facts are trivially landmarks, so they need not be verified. For each of the other candidates $L$, the RPG corresponding to $(A_L, I, G)$ is built, and if that RPG reaches the goals, then $L$ and its incident edges are removed from the LGG.

Reconsider the road map example depicted in Figure 5. The LGG built will be {*at(A)*, *at(E)*, *at(D)*}, {*at(A)* $\rightarrow_{gn}$ *at(E)*, *at(E)* $\rightarrow_{gn}$ *at(D)*}. But *at(E)* is not really a landmark because the action sequence ⟨*move(A,B)*, *move(B,C)*, *move(C,D)*⟩ achieves *at(D)*. When verifying *at(E)*, we detect this. In the RPG, when ignoring all actions that achieve *at(E)*, *move(A,B), move(B,C), and move(C,D)* stay in and so the goal remains reachable. Thus *at(E)* and its edges (in particular, the invalid edge *at(E)* $\rightarrow_{gn}$ *at(D)*) are removed, yielding the final (trivial) LGT with node set {*at(A)*, *at(D)*} and empty edge set. Note that, if *at(E)* was a landmark for some other reason than reaching $D$ (like, if one had to pick up some object at $E$), then *at(E)* would not be removed by landmark verification and the invalid order *at(E)* $\rightarrow_{gn}$ *at(D)* would stay in.

In the *Blocksworld* example from Figure 1, landmark verification does not remove any candidates, and the LGG remains unchanged as depicted in Figure 7.

### 4.2 Approximating Reasonable Orders

Our process to approximate reasonable orders starts from the LGG as computed by the methods described above, and enriches the LGG with new edges corresponding to the approximated reasonable orders. The process has two main aspects:

1. **We approximate the aftermath relation based on the LGG.** This is done by evaluating a sufficient condition that covers certain cases when greedy necessary orders imply the aftermath relation.

2. **We combine the aftermath relation with interference information to approximate reasonable orders.** For each pair of landmarks $L'$ and $L$ such that $L'$ is in the aftermath of $L$ according to the previous approximations, a sufficient condition is evaluated. The condition covers certain cases when $L$ interferes with $L'$, i.e., when achieving $L$ (from a state in $S_{(L',\neg L)}$) involves deleting $L'$. If the condition holds, a reasonable order $L \rightarrow_r L'$ is introduced.

The next two subsections focus on these two aspects in turn. In our implementation, the computation of the aftermath relation is interleaved with its combination with interference information. Pseudo-code for the overall algorithm is given in the second subsection, Figure 8.

#### 4.2.1 Aftermath Relation

The sufficient condition that we use to approximate the aftermath relation is the following.

**Lemma 1** *Given a planning task* $(A, I, G)$*, and two landmarks* $L$ *and* $L'$*. If either*

1. $L' \in G$*, or*

2. *there are landmarks* $L = L_1, \dots, L_{n+1}$, $n \geq 1$, $L_n \neq L'$, *such that* $L_i \rightarrow_{gn} L_{i+1}$ *for* $1 \leq i \leq n$, *and* $L' \rightarrow_{gn} L_{n+1}$,

*then* $L'$ *is in the aftermath of* $L$*.*

**Proof Sketch:** If $L' \in G$ then $L'$ is trivially in the aftermath of $L$. Otherwise, under the given circumstances, $L'$ and $L_n$ must be true together at some point in any action sequence achieving the goal from a state in $S_{(L',\neg L)}$, namely directly prior to achievement of $L_{n+1}$.

As $L$ has a path of $\rightarrow_{gn}$ orders to $L_n$, it has to be true prior to (or simultaneously with, if $n = 1$) $L'$. □

Note that this lemma just captures the property we mentioned before, when we can tell from the LGG that several facts must be true together at some point. In the second case of the lemma, these facts are $L'$ and $L_n$. $L'$ and $L_n$ are both ordered $\rightarrow_{gn}$ before $L_{n+1}$ and so must be true together before achieving that fact. The first case of the lemma can be understood this way as implicitly assuming $L_n$ as some other top-level goal that $L$ has a path of $\rightarrow_{gn}$ orders to. (In our implementation, $L$ must have such a path in the LGG or it would not have been generated as a landmark candidate.) If $L'$ and $L_n$ must be true together, and we additionally know that $L$ must be true sometime before $L_n$, then we know that $L'$ is in the aftermath of $L$.[7]

The most straightforward idea to make use of Lemma 1 would be to simply enumerate all pairs of nodes (landmarks) in the LGG, and evaluate the lemma, collecting the pairs $L$ and $L'$ of landmarks where the lemma condition holds. While this would probably not be prohibitively runtime-costly, one can do better by having a closer look at the lemma condition. Consider each node $L'$ in the LGG in turn. If $L'$ is a top level goal, then $L'$ is in the aftermath of all other nodes $L$. If $L'$ is not a top level goal, then consider all nodes $L_n \neq L'$ such that $L'$ and $L_n$ both have a $\rightarrow_{gn}$ order before some other node $L_{n+1}$. The nodes $L$ in the LGG that have an (possibly empty) outgoing $\rightarrow_{gn}$ path to such an $L_n$ are exactly those for which $L'$ is in the aftermath of $L$ according to Lemma 1. As said, pseudo-code for our overall approximation of reasonable orders is given below in Figure 8.

Note that the inputs to Lemma 1 are $\rightarrow_{gn}$ orders, while in practice we evaluate the lemma on the edges in the LGG as generated by the processes described above in Section 4.1. As we discussed above, the edges in the LGG may be unsound, i.e. they do not provably correspond to $\rightarrow_{gn}$ orders. In effect, neither can we guarantee that our approximation to the aftermath relation is sound.

#### 4.2.2 Reasonable Orders

We approximate reasonable orders by considering all pairs $L$ and $L'$ where $L'$ is in the aftermath of $L$ according to the above approximation. We test if $L$ interferes with $L'$ according to the definition directly below. If the test succeeds, we introduce the order $L \rightarrow_r L'$.

**Definition 6** *Given a planning task $(A, I, G)$, and two facts $L$ and $L'$. $L$* interferes *with $L'$ if one of the following conditions holds:*

1. *$L$ and $L'$ are inconsistent;*
2. *there is a fact $x \in \bigcap_{a \in A, L \in add(a)} add(a)$, $x \neq L$, such that $x$ is inconsistent with $L'$;*
3. *$L' \in \bigcap_{a \in A, L \in add(a)} del(a)$;*
4. *or there is a landmark $x$ inconsistent with $L'$ such that $x \rightarrow_{gn} L$.*

7. In theory, one could also allow $L' = L_n \neq L$ in Lemma 1. In this case, $L$ has a path of $\rightarrow_{gn}$ orders to $L'$, which trivially implies that $L'$ is in the aftermath of $L$. But in fact, it is then impossible to achieve $L'$ before $L$ so an order $L \rightarrow_r L'$ would be meaningless.

As said before, our (standard) definition of inconsistency is that facts $x$ and $y$ are inconsistent in a planning task if there is no reachable state in the task that contains both $x$ and $y$.[8] Note that the conditions 1 to 4 of Definition 6, while they may look closely related at first sight (and presumably are related in many practical examples), indeed cover different cases of when achieving $L$ involves deleting $L'$. More formally expressed, for each condition $i$ there are cases where $i$ holds but no condition $j \neq i$ holds. For example, consider condition 2. In the following example, there is a reasonable order $L \rightarrow_r L'$, and $L$ interferes with $L'$ due to condition 2 only. There are the six facts $L$, $L'$, $x$, $P_1$, $P_2$, and $P'$. Initially only $P'$ is true, and the goal is to have $L$ and $L'$. The actions are:

$$
\begin{array}{lclrr}
\text{name} & & (pre, & add, & del) \\
\mathbf{op}L' & = & (\{P'\}, & \{L'\}, & \{x\}) \\
\mathbf{op}P_1 & = & (\{P'\}, & \{P_1\}, & \{L', P'\}) \\
\mathbf{op}P_2 & = & (\{P'\}, & \{P_2\}, & \{L', P'\}) \\
\mathbf{op}L_1 & = & (\{P_1\}, & \{L, x, P'\}, & \{P_1\}) \\
\mathbf{op}L_2 & = & (\{P_2\}, & \{L, x, P'\}, & \{P_2\})
\end{array}
$$

In this example, the only action sequences that are possible are of the form $(\mathbf{op}L' \parallel \mathbf{op}P_1 \circ \mathbf{op}L_1 \parallel \mathbf{op}P_2 \circ \mathbf{op}L_2)^*$, in BNF-style notation. In effect, $L \rightarrow_r L'$ because if we achieve $L'$ first, we have to apply one of $\mathbf{op}P_1$ and $\mathbf{op}P_2$, which both delete $L'$. Condition 2 holds: $x$ is inconsistent with $L'$ and added by both $\mathbf{op}L_1$ and $\mathbf{op}L_2$. As for condition 1, $L$ and $L'$ are not inconsistent because one can apply $\mathbf{op}L'$ after, e.g., $\mathbf{op}L_1$. Condition 3 is obviously not fulfilled, and condition 4 is not fulfilled because there are two options to achieve $L$ so no fact has a $\rightarrow_{gn}$ order before $L$.

Interference together with the aftermath relation implies reasonable orders between landmarks.

**Theorem 6** *Given a planning task $(A, I, G)$, and two landmarks $L$ and $L'$. If $L$ interferes with $L'$, and either*

1. *$L' \in G$, or*
2. *there are landmarks $L = L_1, \ldots, L_{n+1}$, $n \geq 1$, $L_n \neq L'$, such that $L_i \rightarrow_{gn} L_{i+1}$ for $1 \leq i \leq n$, and $L' \rightarrow_{gn} L_{n+1}$,*

*then there is a reasonable order between $L$ and $L'$, $L \rightarrow_r L'$.*

**Proof Sketch:** By Lemma 1, $L'$ is in the aftermath of $L$. Let us look at the four possible reasons for interference. If $L$ is inconsistent with $L'$ then obviously achieving $L$ involves deleting $L'$. If all actions that achieve $L$ add a fact that is inconsistent with $L'$, the same argument applies. The case where all actions that achieve $L$ delete $L'$ is obvious. As for

8. Deciding about inconsistency is obviously PSPACE-hard. Just imagine a task where we insert one of the facts into the initial state, and the other fact such that it can only be made true once the original goal has been achieved. We approximate inconsistency with a sound but incomplete technique developed by Maria Fox and Derek Long (1998), see below.

the last case, say we are in a state $s \in S_{(L', \neg L)}$. Then $x$ is not in $s$ (because $L'$ is). Due to $x \rightarrow_{gn} L$, $x$ must be achieved directly prior to $L$, and thus $L'$ will be deleted. □

Overall, our method for approximating reasonable orders based on the LGG works as specified in Figure 8. With what was said above, the algorithm should be self-explanatory except for the interference tests. When doing these tests, we need information about fact inconsistencies, and, for condition 4 of Definition 6, about $\rightarrow_{gn}$ orders. Our approximation to the latter pieces of information are, as before, the (approximate) $\rightarrow_{gn}$ edges in the LGG. Our approximation to the former piece of information is a technique from the literature (Fox & Long, 1998), the TIM API. This provides a function *TIMinconsistent(x,y)* that, for facts x and y, returns TRUE only if x and y are inconsistent. The function is incomplete, i.e., it can return FALSE even if x and y are inconsistent.

```
for all nodes L' in the LGG do
    if L' ∈ G then
        for all nodes L ≠ L' in the LGG do
            if L interferes with L', then insert the edge L →r L' into the LGG
        endfor
    else
        for all nodes Ln ≠ L' in the LGG
                s.t. there are a node Ln+1 and edges L' →gn Ln+1, Ln →gn Ln+1 in the LGG do
            for all nodes L in the LGG
                    s.t. L has an (possibly empty) outgoing path of →gn edges to Ln do
                if L interferes with L', then insert the edge L →r L' into the LGG
            endfor
        endfor
    endif
endfor
```

Figure 8: Approximating reasonable orders based on the LGG.

Note that the algorithm from Figure 8 might generate orders $L \rightarrow_r L'$ in cases where $L$ already has a path of $\rightarrow_{gn}$ edges to $L'$. As noted earlier, in this case $L'$ can not be achieved before $L$ so the order $L \rightarrow_r L'$ is meaningless. One could avoid such meaningless orders by an additional check to see, for every generated pair $L$ and $L'$, if $L$ has an outgoing $\rightarrow_{gn}$ path to $L'$. We do this in our implementation only for the easy-to-check special cases where the length of the $\rightarrow_{gn}$ path from $L$ to $L'$ is 1 or 2. Note that the superfluous $\rightarrow_r$ orders don't hurt anyway; in fact they don't change our search process (Section 5.1) at all. The only purpose of our special case test is to avoid some unnecessary evaluations of Definition 6.

Because the inputs to our approximation algorithm are $\rightarrow_{gn}$ edges in the LGG, and as discussed before these edges are not provably sound, the resulting $\rightarrow_r$ orders are not provably sound (which they otherwise would be by Theorem 6).

Let us finish off our running *Blocksworld* example, by showing how the order *clear(C)* $\rightarrow_r$ *on(B D)*, our motivating example from the introduction, is found. Have a look at the LGG in Figure 7. Say the process depicted in Figure 8 considers, in its outermost **for**-loop, the LGG node $L'$ = *on(B D)*. $L'$ is a top level goal so all other nodes $L$ in the LGG, in particular $L$ = *clear(C)*, are considered in the inner **for**-loop. Now, *clear(C)* interferes with

*on(B D)* because of condition 4 in Definition 6: *clear(D)* is inconsistent with *on(B D)*, and has an edge *clear(D)* $\rightarrow_{gn}$ *clear(C)* in the LGG. Consequently the order *clear(C)* $\rightarrow_r$ *on(B D)* is inferred and introduced into the LGG. Note that, to make this inference, we need the edge *clear(D)* $\rightarrow_{gn}$ *clear(C)* which is *not* a $\rightarrow_n$ order.

### 4.3 Approximating Obedient Reasonable Orders

The process that approximates obedient reasonable orders starts from the LGG already containing the approximated reasonable orders, and inserts new orders that are reasonable given one commits to the $\rightarrow_r$ orders already present in the LGG. The technology is very similar to the technology we use to approximate reasonable orders. Largely, we do the same thing as before and just treat the $\rightarrow_r$ edges as if they were additional $\rightarrow_{gn}$ edges. Formally, the difference lies in the sufficient criterion for the, now obedient, aftermath relation.

**Lemma 2** *Given a planning task $(A, I, G)$, a set $O$ of reasonable ordering constraints, and two landmarks $L$ and $L'$. If either*

1. *$L' \in G$, or*
2. *there are landmarks $L = L_1, \ldots, L_{n+1}$, $n \geq 1$, $L_n \neq L'$, such that $L_i \rightarrow_{gn} L_{i+1}$ or $L_i \rightarrow_r L_{i+1} \in O$ for $1 \leq i \leq n$, and $L' \rightarrow_{gn} L_{n+1}$,*

*then $L'$ is in the obedient aftermath of $L$.*

**Proof Sketch:** By a simple modification of the proof to Lemma 1. The first case is obvious, in the second case $L'$ must be true one step before $L_{n+1}$ becomes true, and $L$ must be true sometime before that. □

Note that the proved property does not hold if there is only a reasonable order between $L'$ and $L_{n+1}$, $L' \rightarrow_r L_{n+1}$ instead of $L' \rightarrow_{gn} L_{n+1}$, even if we have committed to obey $L' \rightarrow_r L_{n+1}$. It is essential that $L'$ must be true *directly* before $L_{n+1}$.[9]

The parts of our technology that do not depend on the aftermath relation remain unchanged. Interference is defined exactly as before. Together with the obedient aftermath relation, it implies obedient reasonable orders between landmarks.

**Theorem 7** *Given a planning task $(A, I, G)$, a set $O$ of reasonable ordering constraints, and two landmarks $L$ and $L'$. If $L$ interferes with $L'$, and either*

1. *$L' \in G$, or*
2. *there are landmarks $L = L_1, \ldots, L_{n+1}$, $n \geq 1$, $L_n \neq L'$, such that $L_i \rightarrow_{gn} L_{i+1}$ or $L_i \rightarrow_r L_{i+1} \in O$ for $1 \leq i \leq n$, and $L' \rightarrow_{gn} L_{n+1}$,*

*then there is an obedient reasonable order between $L$ and $L'$, $L \rightarrow_r^O L'$.*

---

9. If obeying an oder $L_1 \rightarrow_r L_2$ is defined to include the case where $L_1$ and $L_2$ are achieved simultaneously, the lemma does not hold. The facts $L_1, \ldots, L_{n+1}$ could then all be achieved with a single action, given the orders between them are all (only) taken from the set $O$. One can "repair" the lemma by requiring that, for at least one of the $i$ where $L_i \not\rightarrow_{gn} L_{i+1}$ but $L_i \rightarrow_r L_{i+1} \in O$, there is no action that has both $L_i$ and $L_{i+1}$ in its add list.

**Proof Sketch:** By Lemma 2 and the same arguments as in the proof to Theorem 6. □

Our overall method for approximating obedient reasonable orders based on the LGG is depicted in Figure 9. Compare the algorithm to the one depicted in Figure 8. Similarly to before, the new process enumerates all fact pairs $L$ and $L'$ where $L'$ is in the obedient aftermath of $L$ according to the $\rightarrow_{gn}$ and $\rightarrow_r$ edges in the LGG, and Lemma 2. Those pairs $L$ and $L'$ where $L'$ is a top level goal are skipped – these pairs have all already been considered by the process from Figure 8. When $L'$ is not a top level goal, the more generous applicability condition of Lemma 2, see the lines marked $(*)$ in Figure 9, may allow us to find more facts $L$ that $L'$ is in the, now obedient, aftermath of. The generated pairs are tested for interference and, if the test succeeds, an order $L \rightarrow_r^O L'$ is introduced. The test for interference is exactly the same as before. The TIM API (Fox & Long, 1998) delivers the inconsistency approximation, and condition 4 of Definition 6 uses the approximated $\rightarrow_{gn}$ orders in the LGG. Note here that the $x \rightarrow_{gn} L$ order in condition 4 of Definition 6 can not be replaced by an $x \rightarrow_r L$ order even if we have committed to obey the latter order. The validity of the condition depends on the fact that, with $x \rightarrow_{gn} L$, $x$ must be true *directly* before $L$.

**for** all nodes $L'$ in the LGG, $L' \notin G$ **do**
    **for** all nodes $L_n \neq L'$ in the LGG
            s.t. there are a node $L_{n+1}$, an edge $L' \rightarrow_{gn} L_{n+1}$, and
            either $L_n \rightarrow_{gn} L_{n+1}$ or $L_n \rightarrow_r L_{n+1}$ in the LGG **do** $(*)$
        **for** all nodes $L$ in the LGG
                s.t. $L$ has an (possibly empty) outgoing path of $\rightarrow_{gn}$ or $\rightarrow_r$ edges to $L_n$ **do** $(*)$
         if $L$ interferes with $L'$, then insert the edge $L \rightarrow_r^O L'$ into the LGG
        **endfor**
    **endfor**
**endfor**

Figure 9: Approximating obedient reasonable orders based on the LGG.

Note that the approximation algorithm for $L \rightarrow_r^O L'$ orders only makes use of the $\rightarrow_{gn}$ and $\rightarrow_r$ edges in the LGG as computed previously, *not* of the newly generated $L \rightarrow_r^O L'$ edges. One could, in principle, allow also these latter edges in the conditions marked $(*)$ in Figure 9, and install a fixpoint loop around the whole algorithm. This way one would generate obedient reasonable orders, and obedient obedient reasonable orders, and so on until a fixpoint occurs. We did not try this, but our intuition is that it typically won't help to improve performance. It seems questionable if, in examples not especially constructed to provoke this, useful orders will come up in fixpoint iterations later than the first one.

Taking the LGG as input, our approximated $\rightarrow_r^O$ orders, like the approximated $\rightarrow_r$ orders, inherit the potential unsoundness of the approximated $\rightarrow_{gn}$ orders.

### 4.4 Cycle Handling

As mentioned earlier, the final LGG including all $\rightarrow_{gn}$, $\rightarrow_r$, and $\rightarrow_r^O$ orders may contain cycles. An example for facts $L$ and $L'$ where both $L \rightarrow_r L'$ and $L' \rightarrow_r L$ hold was given in Section 3.2. Also, cycles $L \rightarrow_{gn} L'$ and $L' \rightarrow_r L$ might arise if a landmark must be achieved more than once in a solution plan (in the *Blocksworld* domain, cycles of this kind sometimes arise for the fact *arm-empty()*). In our current implementation, any cycles in the LGG are removed since the search process can't handle them. The cycles are removed by removing edges – ordering constraints – that participate in cycles. Obviously, one would like to remove as few edges as possible. But figuring out the smallest edge set sufficient to break all cycles is NP-hard (FEEDBACK ARC SET, see Garey & Johnson, 1979). We experimented with a variety of methods that greedily remove edges until there are no more cycles. In these experiments, done over a range of benchmark domains, we generally found that there was little to be gained by the different methods. We settled on the following simple removal scheme: first remove all $\rightarrow_r^O$ edges incident on cycles, then, if any cycles remain, remove all $\rightarrow_r$ edges incident on cycles. After this, the cycles are all removed because $\rightarrow_{gn}$ edges alone can not form cycles due to the way they are computed.

We prioritise to keep (greedy) necessary over reasonable over obedient reasonable ordering constraints in the LGG. This makes intuitive sense due to the stronger theoretical justification of the different types of orders. Of course, there are other methods one can try to treat cycles. One possible approach (suggested to us by one of the anonymous reviewers) would be to collapse all the cycles, i.e., to compute the acyclic directed graph of the strongly connected components of the LGG, and say that $L \rightarrow L'$ iff there is a path from $L$ to $L'$ in that graph of components. One could then use this "meta"-order as input to the search control. Another idea would be to try to analyse the cycles for useful search information, or at least for hints as to what edges may be best to remove. Exploring such alternative approaches is an open research topic.

### 4.5 Lookahead Orders

We have implemented another form of orders, to overcome certain shortcomings of the technology described so far. As said, our approximation of (greedy) necessary orders is based on intersecting action preconditions. We get an order $L \rightarrow_{gn} L'$ if all achievers (in the earliest RPG level) of $L'$ share the precondition $L$. Now, there are situations where the immediate achievers of $L'$ do not have a shared precondition, but the achievers that are (at least) *two* steps away have one. We observed this in the *Logistics* domain. Let us use this domain for illustration. Say we are facing a *Logistics* task where there are two planes and a package to be moved from *la-post-office* to *boston-post-office*. While extracting landmarks, we will find that $L' =$ *at(pack1 boston-airport)* is a landmark (a landmark candidate, at this point in the algorithm). $L'$ is achieved by the two actions $a_1 =$ *unload(pack1 plane1 boston-airport)* and $a_2 =$ *unload(pack1 plane2 boston-airport)*. The preconditions of $a_1$ are *at(plane1 boston-airport)* and *in(pack1 plane1)*, those of $a_2$ are *at(plane2 boston-airport)* and *in(pack1 plane2)*. The intersection of these preconditions is empty, so the landmark extraction process described above would stop right here. But no matter if we use $a_1$ or $a_2$ to achieve $L'$, the package has to be *in* a plane beforehand; and all the earliest actions in the RPG that achieve such an *in* relation share the precondition $L =$ *at(pack1 la-airport)*. (The actions are *load(pack1 plane1 la-airport)* and *load(pack1 plane2 la-airport)*). Thus *at(pack1 la-airport)* is a landmark, which can't be found by our above approximation techniques.

We designed a preliminary technique implementing a limited lookahead, during the landmarks extraction process, in order to overcome the difficulty exemplified above. The general situation where we can introduce a *lookahead necessary* order $L \rightarrow_{ln} L'$ between facts $L$ and $L'$ is this. Say $L'$ is a landmark, and can be achieved by the actions $a_1, \ldots, a_n$. Say $\{L_1, \ldots, L_m\}$ is a set of facts such that the precondition of each action $a_i$, $1 \leq i \leq n$, contains at least one fact $L_j$, $1 \leq j \leq m$. Then the disjunction $L_1 \vee \ldots \vee L_m$ is a landmark in the sense that one of these facts must be true at some point on every solution plan. Now, if all actions that achieve any of the facts $L_j$ share some precondition fact $L$, then it follows that $L$ is a landmark – a landmark that has to be true (at least) two steps before $L'$ becomes true. When intersecting preconditions not over all actions but only over the earliest achievers in the RPG, we can apply this general principle to the *Logistics* example above: $L'$ is *at(pack1 boston-airport)*, $L_1$ and $L_2$ are *in(pack1 plane1)* and *in(pack1 plane2)*, and $L$ is *at(pack1 la-airport)*.

The idea is to test, given a landmark candidate $L'$, if there is an intermediate fact set $\{L_1, \ldots, L_m\}$ as above such that the intersection of the preconditions of all achievers of facts in the set is non-empty. Candidates for such intermediate fact sets can be generated by selecting one precondition from each achiever of $L'$. The obvious difficulty is that the number of such candidates is exponential in the number of different achievers of $L'$. We implemented an incomplete solution by restricting the test to candidate sets where all facts are based on the same predicate – as the *in* predicate in *in(pack1 plane1)* and *in(pack1 plane2)* above. If there is such a fact set $\{L_1, \ldots, L_m\}$, then we check if there is a shared precondition $L$ of the actions achieving the facts $L_j$ at the respective earliest points in the RPG. If the test succeeds, $L$ becomes a new landmark candidate. If, during landmark verification, both $L$ and $L'$ turn out to really be landmarks, then we introduce the order $L \rightarrow_{ln} L'$ into the LGT.

When approximating reasonable or obedient reasonable orders, the $\rightarrow_{ln}$ orders are only used at those points where the aftermath relation is checked. More precisely, the second case of Lemma 1 can be updated to say:

> If there are landmarks $L = L_1, \ldots, L_{n+1}$, $n \geq 1$, $L_n \neq L'$, such that $L_i \rightarrow_{gn} L_{i+1}$ or $L_i \rightarrow_{ln} L_{i+1}$ for $1 \leq i \leq n$, and $L' \rightarrow_{gn} L_{n+1}$, then $L'$ is in the aftermath of $L$.

The second case of Lemma 2 can be updated to say:

> If there are landmarks $L = L_1, \ldots, L_{n+1}$, $n \geq 1$, $L_n \neq L'$, such that $L_i \rightarrow_{gn} L_{i+1}$ or $L_i \rightarrow_r L_{i+1} \in O$ or $L_i \rightarrow_{ln} L_{i+1}$ for $1 \leq i \leq n$, and $L' \rightarrow_{gn} L_{n+1}$, then $L'$ is in the obedient aftermath of $L$.

The approximation algorithms from Figures 8 and 9 are updated accordingly, by allowing $\rightarrow_{ln}$ edges in the respective conditions (in Figure 9, marked with $(*)$). Note that Lemmas 1 and 2 do *not* remain valid when allowing $L' \rightarrow_{gn} L_{n+1}$ to be a $\rightarrow_{ln}$ order, because it is important that $L'$ must be true *directly* before $L_{n+1}$.

We found our implementation of this preliminary technique to work well in the *Logistics* domain, see also Section 6.8. From a more general point of view, the technique opens up two interesting lines of future research, regarding *disjunctive landmarks* and *k-lookaheads.* We say more on these topics in the discussion, Section 7.

## 5. How to Use Ordered Landmarks

Having settled on algorithms for computing the LGG, there is still the question of how to use this information to speed up planning. Porteous and Sebastia (2000) proposed a method that forces the planner to obey all constraints in the LGG. The method is applicable (only) in forward state space search. By a *leaf* in a directed graph we mean, in what follows, a node that has no incoming edges. In Porteous and Sebastia's approach, if applying an action achieves a landmark $L$ that is not a leaf of the current LGG, then they disallow that action. If an action achieves a landmark $L$ that *is* a leaf, then they remove $L$ (and all ordering relations it participates in) from the LGG. In short, they do not allow achieving a landmark unless all of its predecessors have been achieved already. Note that the approach assumes, like the approach we will describe below, that there are no cycles in the LGG – otherwise, there are nodes in the LGG that will never become leaves.

Apart from its restriction to a forward search, the performance improvements obtainable by Porteous and Sebastia's approach appear rather limited. In many domains there are no improvements at all.[10] Here, we explore an alternative idea that uses landmarks in a more constructive manner, by ways of a decomposition method. Rather than telling the planner what is probably *not* a good thing to do next, the landmarks now tell the planner what probably *is* a good thing to do next. This yields performance improvements in domains that were previously unaffected, and is not restricted to any particular kind of planning approach.

Section 5.1 introduces our decomposition method. Section 5.2 makes some remarks about theoretical properties of the method, and Section 5.3 discusses a few variations of the method that we have tried (and found to not work as well in practice).

### 5.1 Disjunctive Search Control

We use the LGG to decompose the planning task into a series of smaller sub-tasks. The decomposition takes place in the form of a search control algorithm that is wrapped around some – any – planning algorithm, called *base planner* in what follows. Similar to before (Porteous & Sebastia, 2000), the core technique is to consider the leaf nodes in an incrementally updated LGG. First, read in the task and compute the LGG. Then, in any search iteration, consider the leaf nodes of the current LGG and *hand these leaves over to the base planner as a goal.* When the base planner has returned a plan, update the LGG by removing the achieved leaf nodes, and iterate. The main difficulty with this idea is that the leaf nodes of the LGG can often not be achieved as a conjunction. Our solution is to pose the leaf nodes as a *disjunctive* goal instead. See the algorithm in Figure 10.

The depicted algorithm keeps track of the current state $s$, the current plan prefix $P$, and the current disjunctive goal $Disj$, which is always made up out of the current leaf nodes of the LGG.[11] The initial facts are immediately removed because they are true anyway. When

10. By this kind of search control one does not get any benefit out of the necessary orders, as these will be obeyed by the forward search anyway. But in many domains (like *Logistics*-type problems) there are no or hardly no ordering constraints other than the necessary ones.

11. Note that, in leaf removal, the plan fragment $P'$ is not processed sequentially, i.e. it is not checked what new LGG nodes become leaves as one applies the actions. While one could of course do the latter, this does not seem to be relevant in practice. We never observed plan fragments that achieved LGG nodes that were not yet leaves.

$s := I,\ P := \langle\, \rangle$
remove from LGG all initial facts and their edges
**repeat**
  $Disj$ := leaf nodes of LGG
  call base planner with actions $A$, initial state $s$ and goal condition $\bigvee Disj$
  **if** base planner did not find a solution $P'$ **then** fail **endif**
  $P := P \circ P',\ s :=$ result of executing $P'$ in $s$
  remove from LGG all $L \in Disj$ with $L \in add(o)$ for some $o$ in $P'$
**until** LGG is empty
call base planner with actions $A$, initial state $s$ and goal $\bigwedge G$
**if** base planner did not find a solution $P'$ **then** fail **endif**
$P := P \circ P'$, output $P$

Figure 10: Disjunctive search control algorithm for a planning task $(A, I, G)$, repeatedly calling an arbitrary planner on a small sub-task.

the LGG is empty – all landmarks have been processed – then the algorithm stops, and calls the underlying base planner from the current state with the original (top level) goals. The search control fails if at some point the base planner did not find a solution. A disjunctive goal can be simulated by using an artificial new fact $G$ as the goal, and adding one action for each disjunct $L$, where the action's precondition is $\{L\}$ and the add list is $\{G\}$ (this was first described by Gazen & Knoblock, 1997). So our search control can be used around any base planner capable of dealing with STRIPS input. Note that, as all top-level goals are landmarks, an empty LGG means that all goals have been achieved at least once. So unless they have been destroyed again, the initial state for the last call of the base planner will already contain the goals (some more on this below).

Note that the search control is completely unaware of destructive interactions between different parts of the overall planning task. It may be, for example, that a landmark $L_1$ is needed for (necessarily ordered before) another landmark $L'_1$, that $L_2$ is needed for $L'_2$, that $L_1$ and $L_2$ are both leaves, and that $L_1$ and $L_2$ are inconsistent. With our search control, the planner will make one of $L_1$ or $L_2$ true. Say $L_1$ is made true. Then the planner is free to choose if, in the next iteration, it wants to make $L'_1$ true, or $L_2$ instead. If the choice is to make $L_2$ true, then $L_1$ – whose purpose was to enable achievement of $L'_1$ – will be invalidated and the effort spent in achieving it will (may) be wasted. On a higher level of abstraction, this means that, with the purely disjunctive search control, the planner may be tempted to frequently "switch" between different parts of the task/of the LGG, which may not be beneficial for the overall performance. We observed such a phenomenon in the *Grid* domain, see Section 6.7; it is unclear to us how much effect on performance such phenomena have in our other experimental domains/if they have any effect at all. Probably there is such an effect in *Blocksworld*, *Depots*, *Freecell*, and *Rovers*, but not in the rest of the domains we tried, i.e. *Logistics*, *Tyreworld*, and various domains where not many landmarks/orders were found or where the orders did not have much effect on performance anyway. It is an open research topic to better investigate the empirical effects of conflicting parts of the task/the LGG, and particularly to come up with general definitions and algorithms to capture and utilise the nature of such phenomena. While this may look simple in the situation outlined above, the situations occuring in reality or even in simple benchmark domains such as the Blocksworld are probably much more complex and difficult to reason about.

### 5.2 Theoretical Properties

The search control is obviously correctness preserving – eventually, the planner is run on the original goal. Likewise obviously, the method is not optimality preserving in general. Though we did not explore this topic in depth, we do not believe that there are interesting special cases in which optimality is preserved. In the benchmarks we tried, the runtime improvements were indeed often bought at the price of somewhat longer plans, see Section 6.

With respect to completeness, matters are a little more interesting. Even if the base planner is complete – guarantees to find a plan if the task is solvable – the search control can fail because one of the encountered sub-tasks is unsolvable. Now, one can of course try to rescue completeness (as much as possible) by an appropriate reaction if the search control loop, as depicted in Figure 10, fails. One can, for example, run the base planner on the original initial state and goal in this case. While this would trivially give us completeness, it would be more desirable to make some use of the information obtained in the search process so far. An idea, which we refer to as the *safety net* in what follows, is to call the base planner in case of failure, with the original goal, but with the start state $s$ of the failed iteration as the initial state. The unsolvability of the failed sub-task might be due to the goal condition as given by the disjunction of the current leaf landmarks. Of course, the unsolvability of the failed sub-task can also be due to the state $s$. If the latter can't happen in the task at hand, then the safety net solution is completeness-preserving. A little more formally, we say that a state is a *dead end* if the (original) goal can not be reached from it. We call a task *dead-end free* if there are no dead ends in its state space. Obviously, if one calls a complete base planner with the original goal in case of failure, then in a dead-end free task that is guaranteed to find a plan.

While the safety net solution preserves completeness in the absence of dead ends, its practical value is unclear, even in dead-end free tasks. The only way the search control can possibly speed up the planning process is if we get close enough to a goal state before the base planner has to be called with the original goal. If the whole LGG is processed before that happens, i.e., if the search control as depicted in Figure 10 does not fail, then it is reasonable to assume that the final state $s$ will be close to the goal – all goal facts have been achieved at least once on the path to $s$. If, however, the control fails before the LGG is empty, then it is completely open how much progress we made. If we fail early on in the search process, then it is likely that the last start state $s$ is not far away from the original initial state. The effort invested into creating the LGG, and into solving the first few sub-tasks, was then in vain.

The relevant remaining theoretical question thus is, *are there interesting special cases where we can be sure to reach the end of the control loop without failing?* The answer is, yes there are. We need a notation. A fact $L$ is called *recoverable* if, when $s$ is a reachable state with $L \in s$, and $s'$ with $L \notin s'$ is reachable from $s$, then a state $s''$ is reachable from $s'$ with $L \in s''$.

**Theorem 8** *Given a solvable planning task $(A, I, G)$, and an LGG where each fact $L$ in the tree is a landmark such that $L \notin I$. If the task is dead-end free, and for all facts $L'$ in the tree it holds that either $L'$ is recoverable, or all orders $L \rightarrow L'$ in the tree are necessary, then running any complete planner within the search control defined by Figure 10 will process the entire LGG without failing.*

**Proof:** Assume that the search control fails at some point before processing the entire LGG. We prove that then, in contradiction to the assumption, the current start state $s$ is a dead end. Because the base planner is complete, we know that the disjunction of all current leaf nodes is unsolvable starting from $s$. Say $L'$ is such a leaf. Say $P'$ is a plan solving the original goal from $s$. $P'$ does not add $L'$, as it would otherwise solve the disjunction. The concatenation of the current prefix $P$ with $P'$ is a solution plan, and $L'$ is a landmark not contained in the initial state, so $P$ adds $L'$. If $L'$ is recoverable, we can achieve it from $s$ and have a contradiction to the unsolvability of the disjunction. Therefore, $L'$ is not recoverable, implying by prerequisite that all orders $L \rightarrow L'$ in the tree are necessary. Say $P = \langle a_1, \ldots, a_n \rangle$, and $L'$ is first added by $a_i$. At this point, $L'$ was not in the disjunction, as it would otherwise have been removed from the tree. So there is some $L$ with $L \rightarrow L'$ that is first added by some action between $a_i$ and $a_n$. But then, $P$ does not obey the ordering constraint $L \rightarrow L'$, which is a contradiction to $L \rightarrow_n L'$. It follows that $s$ is a dead end. □

Verifying landmarks with Proposition 1 ensures that all facts in the LGG really are landmarks. The initial facts are removed before search begins. Many of the current planning benchmarks, for example *Blocksworld*, *Logistics*, *Gripper*, *Hanoi*, *Depots*, and *Driverlog*, are invertible in the sense that every action $a$ has a counterpart $\overline{a}$ that undoes $a$'s effects. Such tasks are dead-end free, and all facts in such tasks are recoverable. An example of a dead-end free domain with only necessary orders is *Simple-Tsp*. Examples of dead-end free domains where non-necessary orders apply only to recoverable facts are *Miconic*-STRIPS and *Grid*. All those domains (or rather, all tasks in those domains) fulfill the requirements for Theorem 8. Note that, in cases where the theorem applies, the search control as depicted in Figure 10 is guaranteed to find a plan if the base planner is complete.

In our experiments, we ran the search control without a safety net. Our reasons were the following. First, if the search control failed, it typically did so very early. Second, we rarely observed a case where failure was due to unsolvability of the leaf landmarks disjunction. Most of the time, the start state of the failed iteration was a dead end. Finally, in our experiments it happened very seldom that the search control failed. The *Freecell* domain was the only domain where we observed that the search control failed in cases where the base planner without the control managed to find a plan. So the issue of what to do in case of failure didn't seem to be very relevant.

### 5.3 Search Control Variations

The disjunctive search control in Figure 10 calls the base planner with the original goal once the whole LGG has been processed. Our (only) hope is that the initial state $s$ for that last call of the base planner is much closer to a goal state than the original initial state was. Recall that all top level goals are achieved at least once on the path to $s$. They thus will be true in $s$ unless they have been deleted again in some later control loop iteration. An obvious idea to avoid the latter phenomenon is, once a top level goal $G$ has been achieved, to force the base planner to keep $G$ true throughout the rest of the control process. This can be done by keeping a *conjunctive goal* $Conj$ in addition to the disjunctive goal $Disj$. $Conj$ then always contains the top level goals that have been achieved so far – that have been leaf landmarks and were removed – and the goal for the base planner is always $Conj \wedge \bigvee Disj$. The problem with this idea is that one or a set of already achieved original goals might be inconsistent with the leaf landmarks in a later iteration. Forcing the achieved goals to be true together with the disjunction yields in this case an unsolvable sub-task, making the control algorithm fail. We observed this in various benchmarks. We tried a number of ideas based on detecting inconsistencies (with the TIM API) between $Conj$ and $Disj$, and removing facts from $Conj$ that participate in an inconsistency. However, this did not help much to avoid unsolvable sub-tasks, due to the incompleteness of TIM's inconsistency approximation, and due to the fact that only *pairwise* inconsistencies are detected. We observed cases where the smallest unsolvable sub-conjunction of landmarks had size 3. Apart from all this, even in cases where forcing truth of achieved top level goals did not make the search control fail, the technique did not yield better runtime or solution length behaviour in our experiments. This is probably due to the fact that, unless such goals are inconsistent with the landmarks ahead, they are kept true anyway.

We remark that, if $G$ is a top level goal that is inconsistent with a landmark $L$, then our approximation techniques introduce the order $L \rightarrow_r G$ by Theorem 6, if the inconsistency is detected. If all these edges $L \rightarrow_r G$ are present in the LGG, $G$ can not become a leaf landmark before any $L$ it is inconsistent with, so once achieved as a leaf landmark $G$ can not be inconsistent with any leaf landmark $L$ in a later iteration. The problems are that non-binary inconsistencies can occur, that TIM's approximation of binary inconsistencies is incomplete, and that edges $L \rightarrow_r G$ might be removed during our (uninformed) removal of cycles.

There is another aspect of the search control defined by Figure 10 that, at first sight, appears to be an obvious shortcoming: while the leaf landmarks can often not be achieved together in one piece, this does not justify the assumption that *all single leaves must be achieved separately.* But posing the leaves as a disjunctive goal suggests the latter to the base planner. In some domains (like *Logistics*), this increases plan length considerably (see also Section 6.8). A more precise way of dealing with inconsistencies among leaf landmarks is to hand the leaves over to the base planner in the form of a DNF goal, where the members of the disjunction are a partition of the leaf landmarks into *maximal consistent subsets.* Similar to what we have seen above, an important difficulty with this idea is that inconsistency can not be determined exactly (without solving the planning task in the first place). One can approximate consistency by pairwise consistency according to the TIM API, and obtain a partition into maximal consistent subsets in a greedy manner.[12] But of course undetected or non-binary inconsistencies can make the search control fail. We

12. Finding a partition (DNF) with minimal number of subsets (disjuncts) is NP-hard: this solves PARTITION INTO CLIQUES (Garey & Johnson, 1979) when the graph edges are the pairs of facts that are consistent.

observed this in various benchmarks. Moreover, in our experiments, independently of the base planner used, the effect of the modified technique on performance was not convincing even in those cases were it preserved solvability. There are a few domains (e.g. *Logistics*) were, compared to the fully disjunctive control, the technique improved plan length at the cost of longer runtimes. In other domains (e.g. *Blocksworld-arm*) the modification made no significant difference, in some domains (e.g. *Freecell*) the modification produced clearly worse runtime *and* plan length behaviour.

We remark that, when using both modifications outlined above, our search control becomes a generalisation of the "Goal Agenda Manager" proposed by Koehler and Hoffmann (Koehler, 1998; Koehler & Hoffmann, 2000). In the goal agenda, the top level goals are partitioned into a series of subsets that respects their (approximated) reasonable ordering constraints. The goal handed over to the base planner in any control loop iteration $i$ is then the union (conjunction) of all partition subsets up to point $i$. Restricted to top level goals, the modified disjunctive search control becomes exactly this. This is because the leaf goals in iteration $i$ correspond exactly to the respective partition subset. Consider the workings of the search control under the above modifications. Top level goals are consistent (in solvable tasks) so will be posed as a single conjunctive goal when considering maximal consistent subsets. The previously achieved leaf landmarks will be exactly the goals in previous partition subsets, so when keeping them in the conjunctive goal $Conj$ we end up with exactly what the Goal Agenda Manager does. The generalisation lies in that, in the presence of non-top level goal landmarks, the goal agenda process is enriched with more facts and ordering relations, and a combination of conjunctive and disjunctive (sub-)goals. However, as outlined above, such a control process suffers from difficulties arising from inconsistencies between goal facts and other landmarks, and the simpler fully disjunctive search control framework from Figure 10 typically works better. So we have concentrated on this simple framework in our more extensive experiments.

## 6. Results

We include subsections describing how we set up our suite of testing examples (Section 6.1), describing the planners tested in the experiment (Section 6.2), and describing our results in the eight individual domains used in the tests (Sections 6.3 to 6.10). We finally include a subsection summarizing our observations regarding unsound orders in the LGG (Section 6.11).

### 6.1 Test Suite

To come up with a test suite, we ran preliminary tests in 19 different STRIPS benchmark domains, including all (STRIPS) examples used in the AIPS-1998, AIPS-2000, and AIPS-2002 competitions. We selected 8 out of the 19 domains for more extensive tests. The selected domains are: *Blocksworld-arm*, *Blocksworld-no-arm*, *Depots*, *Freecell*, *Grid*, *Logistics*, *Rovers*, and *Tyreworld*. The reasons why we discarded the other domains were the following. The *Movie* instances were trivial to solve for any planner configuration we tried. In *Driverlog*, *Gripper*, *Mprime*, *Mystery*, *Satellite*, and *Zenotravel*, no (or only very few) non-trivial landmarks/ordering relations were detected so the decomposition imposed by our disjunctive search control essentially came down to serialising the goal set (i.e., achiev-ing all goals one after the other).[13] In *Hanoi* there was a non-trivial LGT but whether to use it or not made no significant difference to any planner's runtime performance (intuitively, because the ordering information is overpowered by the exponentiality in the length of the *Hanoi* solutions). In the STRIPS version of *Schedule*, where operators have a lot of parameters, none of our planners could cope with the pre-processing (namely, the grounding of the operator parameters with all available objects). In *Ferry* and *Miconic-STRIPS*, finally, the performance observations we made were basically the same as those that we made in *Logistics*. As all these three domains are also semantically very similar, we focussed on only a single representative of this class of domains.

For each of the 8 domains we selected for experimentation, we used a problem generator to produce a large suite of test examples. The examples scale in the respective domain's size parameters (e.g., number of blocks in the *Blocksworld* variants), and of each size there are several (in most domains, 20) random instances.

### 6.2 Tested Planners

We chose to run our disjunctive search control around the three planners FFv1.0, FFv2.3, and LPG. The reason for this choice was that we wanted to show the effects of our techniques on the state-of-the-art in sub-optimal planning. (As our techniques do not preserve optimality there is not much point in using them with optimal planners, which are generally outperformed by suboptimal planners anyway.) FFv1.0 (Hoffmann, 2000) is an early STRIPS version of FF that does not use any goal ordering techniques. FFv2.3 (Hoffmann & Nebel, 2001) is the version of FF that participated in the AIPS-2000 and AIPS-2002 planning competitions. The planner enhances FFv1.0 with the Goal Agenda Manager technique (as well as the ability to handle ADL). We consider FFv2.3 a particularly interesting planner to try our landmarks techniques on as, by doing this, we give an example of how our search control affects the performance of a planner that already uses goal ordering techniques. The LPG version we used (Gerevini et al., 2003) is the one that participated in the AIPS-2002 planning competition.

The implementation of our landmarks techniques is based on FFv1.0, and the disjunctive search control is integrated in that planner's code. For the other planners, FFv2.3 and LPG, we implemented a simple interface. For each iteration of the disjunctive search control, the respective sub-task is specified via two files in the STRIPS subset of PDDL (McDermott et al., 1998). The implementations of FFv2.3 and LPG are modified to output a results file containing the spent running time, and a sequential solution plan (or a flag saying that no plan has been found). For FFv1.0, the runtime we measure is simply total execution time. For FFv2.3, the runtime we measure also is total execution time, except the time taken in the interface, i.e. the time taken to create the PDDL files – after all, this time is just an unnecessary overhead due to our preliminary implementation. For LPG, matters are a bit more complicated. LPG computes inconsistent facts and actions as a pre-process to planning (inconsistent actions either interfere, or have inconsistent preconditions). Repeat-

13. In the transportation domains *Driverlog*, *Gripper*, *Mprime*, *Mystery*, and *Zenotravel*, due to the presence of several vehicles (gripper hands in the case of *Gripper*) there are no shared preconditions of the actions that can achieve the goal position of an object. The only ordering relations we get refer to the initial position and the goal position of objects. We elaborate this further in the discussion of *Logistics* in Section 6.8.

edly calling LPG inside our landmarks control results in repeatedly doing the inconsistency pre-process, producing a large runtime overhead. In a direct implementation of landmarks control within LPG, one would of course do the pre-process only once. Our idea for an experimental implementation thus is to simply ignore the runtime overhead incurred by the superfluous pre-processes. However, in general the inconsistencies computed by the individual pre-processes can be different, depending on the start state LPG is run on. So there is no guarantee that our experimental implementation will produce the same results as a direct implementation, *unless* it is the case that LPG's pre-process provably computes the same information throughout our landmarks control loop. But we found that the latter is indeed the case in 7 of our 8 domains. The reason for this is that state reachability is (largely, in some cases) invertible in these domains; started in states that are reachable from each other, LPG's pre-process finds the same inconsistencies. The detailed arguments are in Appendix C. The only domain where LPG's pre-process might find different information for the different start states is *Freecell*. There we count the total runtime of our experimental implementation, including *all* inconsistency pre-processes. In the other 7 domains, we count the runtime for only a single (the first) one of these pre-processes.[14] It is a piece of open work to integrate landmarks control more tightly with LPG, see also the outlook in Section 7. With the preliminary implementation we evaluate here, the LPG results should be interpreted with care; still they show that one can obtain runtime improvements for LPG when using landmarks to structure the search.

We also ran our disjunctive landmarks control around IPP (Koehler, Nebel, Hoffmann, & Dimopoulos, 1997), and a standard naive breadth-first search. In both cases, we obtained dramatic runtime improvements in all the eight selected domains.

In what follows, there are individual subsections for the selected domains, in alphabetical order. In each subsection, we give a brief description of the domain, provide a table with solution percentage values as well as averaged runtimes and plan lengths (number of actions in the plan), and discuss the results. We also discuss what kind of landmarks and orders our approximation methods find in the respective domain, and how this information relates to the orders that really are present in the domain. We deem this information important to understand the practical impact of our approximation techniques. The discussions are not necessary to understand the rest of the paper, and the uninterested reader may skip over them at the end of each subsection.

The presentation of the data in the form of tables was chosen because this is by far the most compact way to present the data gathered in such a large experiment. To foster the understandability of the solution percentage and runtime results, we also provide runtime distribution graphs (number of solved instances plotted against runtime). These are moved into Appendix B since they take a lot of space, and serve here only to clarify a few specific points.

The experiments were run on an Athlon 1800 MHz machine with 1Gb RAM, running Linux (Mandrake 9.0). Unsuccessful runs were cut off after a time limit of 300 seconds.

14. There is a further subtlety regarding another pre-process that LPG performs before starting the search; see the discussion in Appendix C.

| size | 20 | 22 | 24 | 26 | 28 | 30 |
|---|---|---|---|---|---|---|
| **%solved** | | | | | | |
| FFv1.0 | 25 | 30 | 20 | 0 | 5 | 0 |
| FFv1.0+L | 85 | 70 | 65 | 60 | 70 | 70 |
| FFv2.3 | 80 | 75 | 65 | 50 | 55 | 50 |
| FFv2.3+L | 85 | 95 | 80 | 60 | 60 | 65 |
| LPG | 100 | 100 | 90 | 50 | 35 | 30 |
| LPG+L | 100 | 100 | 95 | 85 | 80 | 65 |
| **time** | | | | | | |
| FFv1.0 | 7.33 | 9.28 | 16.00 | - | 46.99 | - |
| FFv1.0+L | 0.29 | 0.16 | 1.05 | 0.34 | 0.26 | 1.01 |
| FFv2.3 | 16.78 | 6.04 | 2.66 | 10.56 | 0.51 | 2.05 |
| FFv2.3+L | 2.17 | 6.41 | 3.54 | 21.88 | 6.38 | 7.08 |
| LPG | 38.60 | 61.58 | 97.44 | 145.20 | 206.48 | 211.43 |
| LPG+L | 9.67 | 24.03 | 63.45 | 31.61 | 99.51 | 112.79 |
| **length** | | | | | | |
| FFv1.0 | 77.20 | 80.50 | 74.67 | - | 92.00 | - |
| FFv1.0+L | 60.00 | 68.50 | 85.33 | 80.83 | 74.00 | 99.57 |
| FFv2.3 | 64.29 | 64.86 | 75.45 | 78.50 | 88.25 | 88.75 |
| FFv2.3+L | 276.43 | 319.00 | 346.55 | 406.50 | 485.25 | 439.75 |
| LPG | 250.70 | 308.90 | 378.33 | 383.00 | 419.71 | 391.33 |
| LPG+L | 219.90 | 255.10 | 313.56 | 317.00 | 338.29 | 456.67 |

Table 1: Experimental Results in *Blocksworld-arm*. Times are in seconds. Time/plan length in each table entry is averaged over the respective instances solved by *both* of each pair "X" and "X+L" of planners. Size parameter is the number of blocks, 20 random instances per size.

### 6.3 Blocksworld-arm

*Blocksworld-arm* is the variant of the Blocksworld that we used in our illustrative example. A robot arm can be used to arrange blocks on a table. There are four operators to stack a block onto another block, to unstack a block from another block, to put a block (that the arm is holding) down onto the table, and to pick a block up from the table. Using the software provided by Slaney and Thiebaux (2001), we generated examples with 20, 22, 24, 26, 28, and 30 blocks, 20 instances per size. Our data is displayed in Table 1.

From left to right, the entries in Table 1 provide the data for the instances of increasing size. For planner "X", "X+L" denotes the same planner, but with our landmarks search control. The top part of the table provides the percentage of instances solved by each planner in each class of instances. The middle part provides averaged runtimes, the bottom part provides averaged plan lengths. For each planner "X", the averages here are computed over (only) those instances that were solved by *both "X" and "X+L"*. In unsolved cases, there is no plan length to include into the average computation; we have tried to use the runtime cutoff value in the runtime average computation but this generally obscured our results more than it helped to understand them. We average over the instances solved by individual planners "X" and "X + L", rather than over instances solved by all planners, because this way we obtain a clearer picture of what the impact of our techniques on each individual planner is. Note that, this way, comparisons between different planners have to be made very carefully. In what follows, we will indeed concentrate on the effects of our landmarks techniques and not say much on inter-planner comparisons.

One might miss information about variance in Table 1, and in all the other tables below. In our experiments, in all cases where planner "X" had a significantly better/worse average runtime/plan length value than planner "X+L", across a set of random instances solved by both planners, in fact "X" was significantly better/worse than "X+L" across *all* the individual instances that were solved by both planners. So the significant runtime/plan length results – those that we will draw conclusions from – are stable across the random elements in our instance generation processes.

Looking at the solution percentage values, it is obvious that the landmarks help to improve all of the tested planners. The improvement is drastic for FFv1.0 and LPG, and less drastic but still significant for FFv2.3. The runtimes get clearly improved for FFv1.0 and LPG; FFv2.3+L is a little slower than FFv2.3 in those cases solved by both planners. FFv1.0 does not solve a single instance with 26 or 30 blocks, and in the respective table entries for FFv1.0+L, we averaged over all instances solved by FFv1.0+L. For FFv1.0 and LPG, the impact on plan length is a bit inconclusive; most of the time the plans become somewhat shorter when using landmarks. A different picture emerges for FFv2.3 where the plans become a lot longer.[15]

There are all kinds of landmarks and ordering relations in *Blocksworld-arm*, as we have already seen during the discussions regarding our illustrative example. No unsound orders are extracted by our approximation methods. There is always just one single action that can achieve a (non-initial) fact for the first time. It is easy to see that this action will also be the first one to achieve the respective fact in the RPG. Therefore, no unsound $\rightarrow_{gn}$ orders will be extracted, c.f. the discussion in Section 4.1.1. It follows that no unsound (obedient) reasonable orders will be extracted either (c.f. Sections 4.2 and 4.3), and that all landmark candidates really are landmarks. Incompleteness of our approximations can arise, e.g., in the approximation of destructive interactions between facts $L$ and $L'$, Definition 6. Say $L$ is a fact *on(A B)*, and $L'$ is a fact *on(C D)* where in the state at hand $D$ is located above $A$ in the same stack of blocks, but $B$ is located somewhere in/on a different stack. Then achieving $L$ involves deleting $L'$ but none of the conditions given in Definition 6 fires.

### 6.4 Blocksworld-no-arm

*Blocksworld-no-arm* is a variant of the Blocksworld where the blocks are moved around directly, i.e. without explicit reference to a robot arm. There are three operators to move a block from another block onto a third block, to move a block from the table onto another block, and to move a block from another block onto the table. As in *Blocksworld-arm*, we used Slaney and Thiebaux's software to generate random examples with 20, 22, 24, 26, 28, and 30 blocks, 20 instances per size. See the data in Table 2.

Solution percentage gets dramatically improved for FFv2.3; for LPG there is some improvement, too, for FFv1.0 the results are a bit inconclusive but of the largest examples, FFv1.0+L solves a lot more than FFv1.0. The average runtimes in solved examples become clearly better for FFv2.3. For FFv1.0 and LPG, they generally become somewhat better too, but not significantly. LPG and LPG+L solved different instances in the largest exam-

15. We suspected that the latter phenomenon is due to an interaction between our landmarks control and the Goal Agenda that FFv2.3 uses. Investigating this, we did however not find such an interaction. Presumably, the odd plan length behaviour of FFv2.3+L is just an effect of minor implementational differences between FFv1.0 and FFv2.3, such as the ordering of facts and actions in the internal representation.

| size | 20 | 22 | 24 | 26 | 28 | 30 |
|---|---|---|---|---|---|---|
| **%solved** | | | | | | |
| FFv1.0 | 95 | 80 | 55 | 85 | 50 | 15 |
| FFv1.0+L | 80 | 70 | 65 | 70 | 35 | 50 |
| FFv2.3 | 90 | 65 | 55 | 60 | 45 | 45 |
| FFv2.3+L | 100 | 95 | 100 | 100 | 95 | 100 |
| LPG | 100 | 100 | 100 | 100 | 80 | 5 |
| LPG+L | 100 | 100 | 100 | 100 | 90 | 20 |
| **time** | | | | | | |
| FFv1.0 | 7.57 | 25.18 | 29.14 | 28.74 | 88.33 | 106.75 |
| FFv1.0+L | 8.93 | 19.89 | 27.36 | 36.34 | 69.84 | 84.32 |
| FFv2.3 | 16.67 | 27.87 | 46.13 | 74.21 | 44.50 | 65.59 |
| FFv2.3+L | 0.86 | 1.78 | 1.78 | 2.34 | 3.22 | 12.11 |
| LPG | 21.71 | 43.19 | 80.45 | 135.58 | 197.32 | - |
| LPG+L | 16.82 | 40.79 | 67.46 | 118.89 | 207.69 | - |
| **length** | | | | | | |
| FFv1.0 | 28.00 | 32.69 | 36.00 | 37.17 | 42.60 | 46.67 |
| FFv1.0+L | 65.20 | 81.38 | 80.12 | 90.92 | 110.00 | 112.67 |
| FFv2.3 | 47.56 | 49.67 | 58.82 | 69.33 | 64.89 | 76.44 |
| FFv2.3+L | 46.67 | 54.67 | 62.18 | 68.17 | 78.44 | 78.78 |
| LPG | 74.00 | 86.85 | 97.45 | 105.05 | 114.00 | - |
| LPG+L | 97.60 | 130.00 | 143.95 | 154.80 | 188.50 | - |

Table 2: Experimental Results in *Blocksworld-no-arm*. Time/plan length in each table entry is averaged over the respective instances solved by *both* of each pair "X" and "X+L" of planners. Size parameter is the number of blocks, 20 random instances per size.

ple group, which is why the respective table entries are empty. As for average plan length, this becomes slightly worse for FFv2.3 and significantly worse for FFv1.0 and LPG.

There are all kinds of landmarks and ordering relations between them in *Blocksworld-no-arm*. For example, in order to move some block $x$ onto a block $y$, $x$ and $y$ must always be clear (so these fact are shared preconditions). There are destructive interactions between the different sub-goals (e.g. a block can not be clear and have some other block on top of it), which lead to reasonable orders. Regarding soundness of our approximation methods in this domain, it is easy to see (from the actions that can be used to achieve a fact for the first time) that the approximations will always be sound. Incompleteness can arise, e.g., due to the same phenomenon as explained for *Blocksworld-arm* above.

While the ordering relations present in *Blocksworld-no-arm* are similar to those in *Blocksworld-arm*, in difference to the latter domain our observed performance improvements are not as significant. One important reason for this is probably that, for all of FFv1.0, FFv2.3, and LPG, *Blocksworld-no-arm* is not as problematic as *Blocksworld-arm* anyway: in both domains our examples contain the same numbers of blocks, but the solution percentages without using landmarks are usually much higher in *Blocksworld-no-arm*.[16]

16. Indeed, Hoffmann (2002) proves that *Blocksworld-no-arm* is an easier domain than *Blocksworld-arm* for planners (such as FF and LPG) using heuristics based on the relaxation that ignores all delete lists.

| size | 20 | 21 | 22 | 23 | 24 | 25 |
|---|---|---|---|---|---|---|
| **%solved** | | | | | | |
| FFv1.0 | 62.5 | 72.5 | 52.5 | 42.5 | 22.5 | 20 |
| FFv1.0+L | 85 | 80 | 75 | 67.5 | 72.5 | 62.5 |
| FFv2.3 | 67.5 | 67.5 | 55 | 45 | 45 | 27.5 |
| FFv2.3+L | 47.5 | 45 | 37.5 | 35 | 30 | 30 |
| LPG | 100 | 100 | 100 | 100 | 100 | 82.5 |
| LPG+L | 100 | 100 | 100 | 95 | 97.5 | 97.5 |
| **time** | | | | | | |
| FFv1.0 | 98.18 | 157.29 | 130.69 | 206.64 | 175.82 | 146.81 |
| FFv1.0+L | 9.48 | 10.32 | 26.47 | 20.12 | 22.33 | 41.01 |
| FFv2.3 | 139.19 | 138.47 | 92.49 | 163.92 | 185.84 | 218.27 |
| FFv2.3+L | 38.75 | 15.10 | 36.27 | 65.76 | 8.74 | 10.63 |
| LPG | 32.39 | 47.59 | 63.89 | 78.72 | 87.23 | 111.45 |
| LPG+L | 22.60 | 26.71 | 46.32 | 51.27 | 53.38 | 74.60 |
| **length** | | | | | | |
| FFv1.0 | 108.86 | 118.00 | 121.00 | 129.64 | 131.00 | 127.43 |
| FFv1.0+L | 128.48 | 146.17 | 152.47 | 154.27 | 170.44 | 161.14 |
| FFv2.3 | 118.42 | 118.00 | 118.00 | 132.38 | 132.20 | 130.00 |
| FFv2.3+L | 133.50 | 141.21 | 150.09 | 159.62 | 154.00 | 159.50 |
| LPG | 137.70 | 155.43 | 157.22 | 165.87 | 178.85 | 184.88 |
| LPG+L | 154.38 | 156.03 | 176.22 | 194.76 | 197.41 | 200.09 |

Table 3: Experimental Results in *Depots*. Time/plan length in each table entry is averaged over the respective instances solved by *both* of each pair "X" and "X+L" of planners. Size parameter is the number of crates; all instances have 10 locations, 3 trucks, 10 hoists, and 20 pallets. 40 random instances per size.

### 6.5 Depots

The *Depots* domain is a mixture between *Blocksworld-arm* and *Logistics*, introduced by Long and Fox in the AIPS-2002 planning competition (Long & Fox, 2003). Stacks of crates at different locations must be re-arranged. There are five operators to drive trucks between places, to load/unload a crate onto/from a truck using a hoist, and to drop/lift a crate onto/from a surface (another crate or a pallet) using a hoist. For our random instances, we used the problem generator provided by Long and Fox. The instances all feature 10 locations, 3 trucks, 10 hoists (one at each location), and 20 pallets (at least one at each location). The instances scale in terms of the number of crates. Here, we show the data for instances with 20 to 25 crates. The runtime of our planners on instances of the same size varied more in *Depots* than in our other domains, and we generated 40 random instances of each size instead of the usual 20. See the data in Table 3.

For all the planners, with our technique the plans become somewhat longer. The solution percentage and runtime values are harder to interpret. Solution percentage for FFv1.0 is drastically improved, while for FFv2.3 it becomes significantly worse. For both planners the average runtime in solved instances becomes a lot better. To understand this, a look at Figure 12 in Appendix B is helpful. This shows quite nicely the "greedy" effect that the landmarks control has on the FF versions in *Depots*. While both FFv1.0 and FFv2.3 solve a lot more examples quickly when using the landmarks control, without that control the planners tend to behave more reliably, i.e. can solve more instances when given enough time. For FFv2.3 this happens before our time cutoff limit.

For solution percentage and runtime of LPG, both the data in Table 3 and the plots in Figure 12 are a bit inconclusive. The solution percentage data suggests that the used instances are not yet at LPG's scaling limit. We generated larger instances, with 26 to 30 crates, 3 random examples of each size. The solution percentage values we got for these examples are (LPG/LPG+L): 100/100, 66.6/100, 0/100, 0/66.6, 0/66.6. This suggests that our landmarks techniques pay off more for LPG as the instances become harder. The runtime and plan length behaviour in cases solved by both planners remained similar to the smaller instances: LPG + L was faster than LPG on average, but found somewhat longer plans.

The hoists and crates in *Depots* roughly correspond to the robot arm and blocks in *Blocksworld-arm*, and similarly to that domain our approximation methods find various landmarks and orders regarding *on* relations, *clear* relations, and, for each location, *lifting* relations for the hoist at that location. The orders $L \rightarrow_{gn} L'$ found by our methods, if $L'$ is an *on* or a *clear* relation, are sound due to the same arguments as given above for *Blocksworld-arm*. This is also true when $L'$ is a *lifting* relation regarding a crate that is initially already co-located with the respective hoist. If the crate is initially not co-located with the hoist, due to the presence of several trucks we get no $\rightarrow_{gn}$ orders: the crate can be unloaded (and is then being lifted afterwards) from any truck, and the unloading actions have no shared preconditions. The only orders our methods find regarding the transportation part of the problem are $L \rightarrow_{ln} L'$ orders, c.f. Section 4.5, where $L'$ is the *lifting* relation for a crate at its goal location, and $L$ is either the crate's *at* relation or its *lifting* relation at its initial location. These orders are sound, $L$ has to be true earlier than $L'$. The orders are found because the unloading actions for all trucks – the actions that achieve $L'$ – all feature an *in* relation for the crate, and these *in* relations form an intermediate fact set for which all earliest achievers (loading actions, namely) in the RPG share the preconditions $L$.[17] Like we observed for *Blocksworld-arm* and *Blocksworld-no-arm*, incompleteness of our approximations may, e.g., arise from the fact that Definition 6 does not cover all destructive interactions that can arise from the way crates are arranged on stacks.

### 6.6 Freecell

The *Freecell* domain is a STRIPS encoding of the well-known solitaire card game that comes free with Microsoft Windows. The domain was introduced by Fahiem Bacchus in the AIPS-2000 planning competition, and was also included by Long and Fox into the AIPS-2002 planning competition. A brief description of the game is as follows. A set of cards is initially randomly arranged across a number of stacks, and all cards have to be put away on goal stacks in a certain order. There are rules that guide in which ways cards can be moved, and there are a number of "free cells" which can be used for intermediate storage of cards. The domain comprises 10 operators that implement the possible card moves. As said above, in *Freecell* there is no guarantee that our experimental implementation behaves

17. If there was only a single truck, then we would get $L \rightarrow_{gn} L'$ orders where $L'$ is a *lifting* relation and $L$ is an *at* relation of the truck. In effect we would get $L \rightarrow_{gn} L'$ orders where both $L$ and $L'$ are an *at* relation of the truck, $L$ corresponding to the truck's initial location. These latter orders can be unsound, but do not affect the rest of the approximation processes, or the search process. The same phenomenon can occur in all transportation domains, and is discussed in the section about *Logistics* below.

| size | 11 | 12 | 13 | 14 | 15 | 16 |
|---|---|---|---|---|---|---|
| **%solved** | | | | | | |
| FFv1.0 | 95 | 85 | 80 | 50 | 35 | 15 |
| FFv1.0+L | 95 | 65 | 60 | 50 | 10 | 5 |
| FFv2.3 | 100 | 90 | 85 | 60 | 20 | 15 |
| FFv2.3+L | 100 | 70 | 70 | 30 | 20 | 5 |
| LPG | 0 | 0 | 0 | 0 | 0 | 0 |
| LPG+L | 10 | 0 | 0 | 0 | 0 | 0 |
| **time** | | | | | | |
| FFv1.0 | 3.95 | 46.82 | 54.28 | 92.60 | - | 261.65 |
| FFv1.0+L | 1.62 | 2.15 | 3.05 | 5.45 | - | 9.34 |
| FFv2.3 | 6.02 | 24.93 | 62.53 | 66.56 | - | 140.37 |
| FFv2.3+L | 1.80 | 2.41 | 3.31 | 4.45 | - | 8.63 |
| LPG | - | - | - | - | - | - |
| LPG+L | 240.23 | - | - | - | - | - |
| **length** | | | | | | |
| FFv1.0 | 68.00 | 88.00 | 103.00 | 107.25 | - | 125.00 |
| FFv1.0+L | 63.00 | 69.50 | 81.50 | 99.25 | - | 119.00 |
| FFv2.3 | 75.50 | 85.00 | 93.00 | 106.50 | - | 123.00 |
| FFv2.3+L | 60.50 | 67.50 | 82.00 | 98.00 | - | 121.00 |
| LPG | - | - | - | - | - | - |
| LPG+L | 72.50 | - | - | - | - | - |

Table 4: Experimental Results in *Freecell*. Time/plan length in each table entry is averaged over the respective instances solved by *both* of each pair "X" and "X+L" of planners. Size parameter is the number of cards per suit; all instances have 4 different suits, 5 free cells, and 10 possible stacks. 20 random instances per size.

like a direct implementation of landmarks in LPG, so below we provide the total runtime taken. We remark that a direct LPG+L implementation might do better – when counting the time for only a single inconsistencies pre-process, we solved a few more examples within the time limit. To generate our test suite, we used the random generator provided by Long and Fox. Across our instances, we have 4 different suits of cards, 5 free cells, and 10 possible stacks; we scaled the instances by increasing the number of cards in each suit, 20 random instances per size. Table 4 provides our data for the examples with 11 to 16 cards per suit.

For a reason we don't know, LPG behaved very badly in our *Freecell* experiments. It hardly solved any instance, and adding the landmarks control helped in some, but not many, cases. Averaging over these cases, we got the runtime and plan length averages shown in the LPG+L entries. For the FF versions, the impact of our landmarks technique is more interesting. The impact on both versions is similar: less tasks are solved, but the runtime in the solved tasks improves considerably, and the plans become somewhat shorter.[18]

An intuitive explanation for the runtime behavior of FF is the following. When playing *Freecell*, the main difficulty is typically to avoid dead end situations where, due to some wrong card moves in the past, one can not move any card any longer, or make any more progress towards a solution. FF uses a forward hill-climbing style search. Pretty much as a matter of chance, this either gets lucky and solves the task quickly, or it gets stuck in a dead end state. Using the landmarks search control, FF's hill-climbing search becomes even more greedy, being a lot faster in the "good" cases but producing a few more "bad" cases. Now,

18. In the examples with 15 cards per suit, the instances solved by FFv1.0/FFv1.0+L (FFv2.3/FFv2.3+L) were different which is why the respective runtime and plan length table entries are empty.

if FF (without our control) encounters a dead end during hill-climbing, the planner starts from scratch, "falling back" onto a complete best-first search. If our landmarks control ends up in a dead end, no such complete fall back search method is invoked (though one could of course do that, c.f. Section 5.2). The instances that FF's complete search method solves cause most of the difference in the solution percentage values as depicted in Table 4. (The greediness, and failure, of the FF versions with landmarks control can also very nicely be observed in Appendix B, Figure 12.)

In a fashion reminiscent of the Blocksworld-like problems considered above, cards in *Freecell* can be stacked on top each other (under certain additional preconditions regarding their value and their colour), and must be clear in order to be moved. From these structures, we get landmarks regarding *on*-relations and *clear*-relations similar as before. In addition, we get landmarks regarding card values in cases where only one order of values is possible. Most particularly, the goal condition is expressed in terms of *home*-relations for (only) the *topmost* cards in the goal stacks. As the order in which cards can be put onto the goal stack is fixed, for each single card in the goal stacks we get its *home*-relation as a landmark, necessarily ordered before the respective next higher card. We can get unsound $\rightarrow_{gn}$ orders because the RPG is a very crude approximation to the semantics of a *Freecell* game. For example, the earliest action in the RPG that can be used to put a card home might be to move the card to home from its initial position, while really the card has to be stored in a free cell first, and be moved home from there. The unsound $\rightarrow_{gn}$ orders may lead to unsound $\rightarrow_r$ orders. If one skips the RPG level test in Figure 4, i.e. if one generates the LGG using the "safe" approximation of $\rightarrow_n$ orders that simply intersects the preconditions of all achieving actions, then the only landmarks and orders one gets are the *home*-relations of the cards, with their associated ($\rightarrow_n$) orders. Using this smaller LGG to control FF, one gets slightly more reliable behaviour. A few examples are solved that are not solved with the potentially unsound approximations. It seems that the additional orders extracted using the RPG are sometimes too greedy and lead into dead ends.

As one would expect, our approximation methods are also incomplete in *Freecell*. For example, Definition 6 can overlook destructive interactions between facts that arise from the stacking order of cards, just like we observed for the blocks/crates in the domains discussed previously.

### 6.7 Grid

*Grid* is a domain that was introduced by Drew McDermott in the AIPS-1998 planning competition. A robot moves on a 2-dimensional grid of locations. The locations can be locked, in which case the robot can not move onto them. Locked locations can be opened with a key of matching shape. The robot can carry one key at a time, and the task is to transport a subset of the keys to their goal locations. For our test examples, we used the random generator provided on the FF homepage.[19] We scaled the instances in terms of parameters $n$ and $m$. The grid has extension $n \times n$, there are 4 different possible shapes of keys and locks, of each shape there are $m$ keys and $m$ locked locations. Half of the keys must be transported to goal locations. Randomly distributing the initial locations of the keys and the locked locations, the generator sometimes produces unsolvable instances. We filtered

19. At `http://www.informatik.uni-freiburg.de/~hoffmann/ff-domains.html`.

| size | 8/4 | 8/5 | 9/4 | 9/5 | 10/5 | 10/6 |
|---|---|---|---|---|---|---|
| **%solved** | | | | | | |
| FFv1.0 | 100 | 85 | 90 | 65 | 30 | 0 |
| FFv1.0+L | 100 | 100 | 100 | 95 | 85 | 90 |
| FFv2.3 | 80 | 50 | 75 | 60 | 25 | 0 |
| FFv2.3+L | 90 | 95 | 80 | 60 | 50 | 50 |
| LPG | 40 | 15 | 35 | 5 | 0 | 0 |
| LPG+L | 80 | 70 | 70 | 5 | 0 | 0 |
| **time** | | | | | | |
| FFv1.0 | 40.92 | 126.82 | 75.85 | 132.44 | 136.24 | - |
| FFv1.0+L | 4.39 | 9.56 | 13.97 | 10.43 | 31.88 | 84.32 |
| FFv2.3 | 35.17 | 64.05 | 74.35 | 91.58 | 113.35 | - |
| FFv2.3+L | 10.53 | 7.09 | 7.71 | 5.59 | 6.84 | 79.82 |
| LPG | 123.37 | 204.63 | 141.80 | - | - | - |
| LPG+L | 151.03 | 277.42 | 169.52 | - | - | - |
| **length** | | | | | | |
| FFv1.0 | 127.15 | 171.71 | 155.17 | 159.77 | 185.33 | - |
| FFv1.0+L | 131.75 | 166.24 | 153.33 | 153.00 | 176.67 | 251.61 |
| FFv2.3 | 127.13 | 145.60 | 149.42 | 149.00 | 152.00 | - |
| FFv2.3+L | 122.93 | 135.50 | 148.50 | 125.33 | 143.75 | 250.10 |
| LPG | 141.29 | 131.50 | 134.50 | - | - | - |
| LPG+L | 149.14 | 131.00 | 141.50 | - | - | - |

Table 5: Experimental Results in *Grid*. Time/plan length in each table entry is averaged over the respective instances solved by *both* of each pair "X" and "X+L" of planners. Size parameters $n/m$ are the extension of the grid ($n \times n$) and the number $m$ of keys and locks of each shape; all instances have 4 different shapes. 20 random instances per size.

these out by hand (whether a *Grid* instance is solvable or not can be concluded from the output of simple pre-processing routines such as are implemented in FF). Table 5 provides our data for the size values ($n/m$) 8/4, 8/5, 9/4, 9/5, 10/5, 10/6, 20 random instances per size.

The results for the FF versions are excellent. Solution percentage and average runtime are improved significantly. This performance gain is not obtained at the cost of longer plans, rather differently the plans using landmarks control become slightly (but consistently) shorter. For LPG, the solution percentage is also improved a lot, but the time taken to solve the examples solved by both planners is somewhat worse for LPG+L. The runtimes are, in part, close to our time cutoff limit so one might suspect that the picture changes when changing that limit. The runtime distribution graph in Figure 13, Appendix B, shows that this is not the case, and that the number of examples solved by LPG consistently grows faster in runtime when using the landmarks control. LPG's plans become a little longer.

The landmarks and orders our approximation techniques find in *Grid* are the following. The goal is given in the form of a set of *at*-relations of keys. These relations can be achieved using putdown actions, which all share the fact that the robot must be *holding* the respective key. Now, in the RPG the earliest point at which *holding* a key is achieved is by picking the key up at its initial location. So we get the respective landmarks for the location of the robot. In some special cases, e.g. when the goal location of a key is locked, we also get landmarks regarding the opening of locked locations. In these cases, unsound $\rightarrow_{gn}$ orders may be extracted because a location can in general be opened from several locations, only one of which may be the earliest one reached in the RPG. Similarly, when a landmark requires the robot to be at a certain location, that can be reached from several other locations, the RPG may suggest just one of these locations as the predecessor. Such unsound orders suggest to the planner one specific way of moving to/of opening the respective location. The unsound $\rightarrow_{gn}$ orders may lead to unsound $\rightarrow_r$ orders in our other approximation processes.

Controlling FF and LPG with the "safe" LGG, $\rightarrow_n$ orders approximated without using the RPG, we found the following. Only few orders were generated: all one gets are the landmarks and $\rightarrow_n$ orders saying, for each goal location of a key, that the robot has to be at this location at some point, and that one has to be holding this key at some point. Somewhat surprisingly, the runtime performance of FF and LPG became a little better with these rather trivial LGGs, as compared to the richer LGGs containing the RPG-approximated orders. Investigating this, we got the impression that the phenomenon is due to conflicts between the individual sub-tasks/parts of the LGG associated with the individual goal keys. With the richer LGGs, the planners tended to keep switching between the sub-tasks, i.e. they tended to keep changing keys and opening locks, not keeping hold of a key and transporting it to its goal location until very late in the planning process. In the late iterations of the search control, for achieving the actual goals comparatively long plans had to be generated. This did not happen as much with the sparser "safe" LGGs, where the goals became leaf nodes earlier on in the control loop.

Regarding completeness of our ordering approximations in *Grid*, as before Definition 6 is not a necessary condition for destructive interactions between facts. E.g. there may be a reasonable order between the goal locations of two keys, because the one key must be used in order to open the path to the other key's goal location. The simple sufficient conditions listed in Definition 6 do not cover such complex cases.

### 6.8 Logistics

The well-known *Logistics* domain is the classical transportation benchmark in planning. Trucks and planes are the means to transport packages between locations. The locations are grouped together in cities, i.e. a single city can contain several locations. Trucks can only drive within cities, planes fly between different cities. There is one truck in each city. There are 6 operators to drive trucks, to fly planes, to load/unload packages onto/from trucks, and to load/unload packages onto/from planes. To generate our test suite, as in *Grid* we used the random generator provided on the FF homepage. Our instances all have 10 cities with 10 locations each, and 10 planes. We scaled the instances by the number of packages (which must all be transported). Table 6 provides our data for the examples with 40, 42, 44, 46, 48, and 50 packages, 20 random instances per size.

The solution percentage/runtime performance of both FF versions is improved drastically by the landmarks technique, at the cost of longer plans. The runtimes without landmarks are close to the runtime cutoff, suggesting that significantly more examples may be solved when increasing the cutoff. Figure 13 in Appendix B shows that this is not the case. For LPG, the solution percentage data in Table 6 is a bit inconclusive but the average runtime over those instances solved by both LPG and LPG+L is better for LPG+L. Looking at Figure 13 one sees that LPG+L is very quick to solve a lot of the examples, but fails to solve the others even if given a lot of time – in difference to LPG without landmarks control, which consistently manages to solve more and more examples until the time cutoff. It is unclear to us why LPG+L stops to scale at some point. LPG's plans become longer with the landmarks control.

| size | 40 | 42 | 44 | 46 | 48 | 50 |
|---|---|---|---|---|---|---|
| **%solved** | | | | | | |
| FFv1.0 | 15 | 20 | 0 | 0 | 0 | 0 |
| FFv1.0+L | 100 | 100 | 100 | 100 | 100 | 100 |
| FFv2.3 | 20 | 20 | 10 | 15 | 0 | 0 |
| FFv2.3+L | 100 | 100 | 100 | 100 | 100 | 100 |
| LPG | 100 | 80 | 90 | 65 | 55 | 35 |
| LPG+L | 100 | 100 | 80 | 70 | 60 | 35 |
| **time** | | | | | | |
| FFv1.0 | 238.11 | 267.20 | - | - | - | - |
| FFv1.0+L | 25.84 | 28.49 | 32.35 | 35.39 | 42.39 | 46.55 |
| FFv2.3 | 207.11 | 233.54 | 226.60 | 228.82 | - | - |
| FFv2.3+L | 58.03 | 63.10 | 69.79 | 76.21 | 91.20 | 112.21 |
| LPG | 124.05 | 170.75 | 189.95 | 204.73 | 246.80 | 202.65 |
| LPG+L | 49.15 | 53.71 | 59.63 | 65.06 | 73.65 | 79.29 |
| **length** | | | | | | |
| FFv1.0 | 351.00 | 351.00 | - | - | - | - |
| FFv1.0+L | 429.00 | 439.00 | 466.60 | 484.65 | 504.75 | 523.90 |
| FFv2.3 | 339.25 | 368.00 | 377.00 | 385.67 | - | - |
| FFv2.3+L | 424.75 | 449.75 | 469.00 | 480.67 | 506.50 | 526.50 |
| LPG | 447.80 | 468.31 | 483.50 | 500.00 | 517.88 | 547.50 |
| LPG+L | 468.95 | 484.38 | 508.36 | 522.89 | 556.88 | 568.00 |

Table 6: Experimental Results in *Logistics*. Time/plan length in each table entry is averaged over the respective instances solved by *both* of each pair "X" and "X+L" of planners. Size parameter is the number of packages; all instances have 10 airplanes, and 10 cities with 10 locations each. 20 random instances per size.

Let us take some time to explain the landmarks and orders found in *Logistics*, in comparison to other transportation domains. The goal in transportation domains is always a set of *at*-relations for transportable objects (sometimes also for vehicles). To achieve the *at* goal for an object, we must unload it from some vehicle, for which the object must be *in* the vehicle. If there is only a single vehicle, like in *Ferry* and *Miconic-STRIPS*, then the *in*-relation for the object is a shared precondition and is thus detected as a landmark, with a necessary order before the *at*-goal. If there are several vehicles, like in *Driverlog*, *Logistics*, *Mprime*, *Mystery*, and *Zenotravel*, then our lookahead technique, c.f. Section 4.5, comes into the play: there is no shared precondition to the unload actions, but the *in*-relations for all vehicles form an intermediate fact set for which all earliest achievers (loading actions, namely) in the RPG share the "earliest loading-*at*-relation of the object". Now, for all the listed domains except *Logistics* the earliest possible loading-*at*-relation of the object simply is its initial location; in *Logistics*, for actions that load the object onto a plane, the earliest possible loading-*at*-relation is the airport of the object's (package's) origin city. So in *Logistics* we get a useful $\rightarrow_{ln}$ lookahead order between the object being *at* its origin and its destination *airport*, while in the other cases we get a non-useful lookahead order between the object being *at* its initial and goal *location*. (A similar observation can be made in *Gripper* where there are several gripper hands.) Overall, for each package in *Logistics* we get an order sequence of the form *at(p initial-location)* $\rightarrow_{gn}$ *in(p origin-city-truck)* $\rightarrow_{gn}$ *at(p origin-city-airport)* $\rightarrow_{ln}$ *at(p destination-city-airport)* $\rightarrow_{gn}$ *in(p destination-city-truck)* $\rightarrow_{gn}$ *at(p goal-location)*.[20] With our disjunctive landmarks control, the decomposition we get divides the package transportation into very small steps and thus helps to improve runtime. The point where we get longer plans is probably the $\rightarrow_{ln}$ order between the package's origin and destination airports: posing the *at*-relations at destination airports as a disjunctive goal, the planner transports the packages between cities one-by-one, without trying to use the same plane for several packages. (As we indicated in Section 5.3, this increase in plan length can be avoided, at the cost of longer runtimes, by posing the leaf landmarks as a DNF goal of maximal consistent subsets – all *at*-relations at destination airports are consistent and the DNF we get is a single conjunction.)

In all transportation domains where the maps on which vehicles move are fully connected, the following phenomenon can occur. Say $v$ is a vehicle, $l$ is its initial location, and $l'$ is some other location. When *at(v l')* becomes a landmark candidate, the order *at(v l)* $\rightarrow_{gn}$ *at(v l')* is generated. But this order is unsound if there is at least one third location $l''$ on $v$'s map.[21] Note, however, that *at(v l)* is true in the initial state so the unsound order will be removed before search begins. It is also easy to see that, in the transportation domains we considered (including *Depots*), the only new orders that the unsound order *at(v l)* $\rightarrow_{gn}$ *at(v l')* can possibly lead to in the rest of our approximation processes are of the form *at(v l)* $\rightarrow_{r}$ $L'$, and will be removed prior to search. We remark that, when using the "safe" LGG without RPG approximation, the only landmarks and orders one gets in *Logistics* say, for each package goal location, that the respective truck has to be at this location at some point, and that the package has to be in this truck at some point. The resulting "decomposition" makes the base planner start by, uselessly, moving all trucks to the package goal locations, and then transport all the packages one-by-one. In comparison to the RPG-approximated richer LGGs, this yields longer plans, and a little more runtime taken by the base planner. (The runtime performance decrease in the base planner is only slight because FF and LPG are very efficient in transporting single packages to their goal.)

Our approximations are complete in *Logistics* in the sense that the only valid orders overlooked by them are reasonable orders between facts $L$ and $L'$ where there is a path of greedy necessary (or lookahead) orders from $L$ to $L'$ anyway. Obviously, our methods find all landmarks in *Logistics*, and all greedy necessary (as well as lookahead) orders between them. Reasonable orders between landmarks in *Logistics* arise only between landmarks $L$ and $L'$ both stating the position of one package/truck/airplane, where the fact interference responsible for the order is the inconsistency of different positions of the package/truck/airplane. Now, TIM detects all these inconsistencies, making condition (1) of Definition 6 a complete approximation of the relevant fact interferences in *Logistics*. In reasonable orders $L \rightarrow_{r} L'$ between positions of one truck/airplane, $L'$ is in the aftermath of $L$ due to situations that are uncovered by the greedy necessary (or lookahead) orders, and Lemma 1. E.g., reasonable orders between truck positions only arise due to situations like, *at(origin-city-truck*

20. Note here that there is just one truck in each city in the standard *Logistics* domain. With several trucks in a single city, we would get a sequence of $\rightarrow_{ln}$ orders between the package's initial location, origin city airport, destination city airport, and goal location.

21. In domains with explicit road map connections, like the *Grid* domain discussed above, the order is unsound if $l'$ can be reached from more than one location on $v$'s road map.

*p-initial-location)* $\rightarrow_r$ *at(origin-city-truck origin-city-airport)* because *at(origin-city-truck p-initial-location)* $\rightarrow_n$ *in(p origin-city-truck)* $\rightarrow_n$ *at(p origin-city-airport)* and *at(origin-city-truck origin-city-airport)* $\rightarrow_n$ *at(p origin-city-airport)*.[22] The only reasonable orders we overlook are those that refer to different positions of a single package on its path *at(p initial-location)* $\rightarrow_{gn}$ *in(p origin-city-truck)* ... *in(p destination-city-truck)* $\rightarrow_{gn}$ *at(p goal-location)*. Later facts on such a path are in the aftermath of the earlier facts, and there are reasonable orders because all the facts are inconsistent. We do not find these orders because we skip (some of the) pairs of facts $L$ and $L'$ where there is a path of greedy necessary (or lookahead) orders from $L$ to $L'$ anyway, c.f. Section 4.2.

### 6.9 Rovers

The *Rovers* domain was introduced by Long and Fox in the AIPS-2002 planning competition. The problem is to find a plan for an interplanetary rovers mission. A set of rovers can navigate along road maps of way points (i.e. the viable road map can be different for different rovers), take soil or rock samples, take images, and communicate their observations back to landers. The goal is to gather together a set of observations. There are 9 operators to navigate the rovers, to take soil/rock samples, to drop such samples (they occupy storage space), to calibrate cameras, to take images, and to communicate soil/rock/image data to landers. We generated our random instances with the software provided by Long and Fox. This software takes as inputs the number of rovers, the number of waypoints, the number of cameras, the number of "modes" (different types of images that can be taken), and the number of objectives (of which images can be taken). We generated instances featuring only a single rover, because we found that, with several rovers, similarly to what we observed in the transportation domains *Driverlog*, *Mprime*, *Mystery*, and *Zenotravel*, no non-trivial landmarks were detected (see also below). We scaled the instances via a parameter $n$. Each instance has one rover, $40+n$ waypoints, $30+n$ cameras, 5 modes, and $n$ objectives. Table 7 shows our data for the $n$ values 30, 32, 34, 36, 38, and 40, 20 random instances per size.

The solution percentage/runtime performance of both FF versions improves dramatically when using our landmarks control. Figure 13 in Appendix B shows that the solution percentage values for the FF versions without landmarks control would not change much with a higher runtime cutoff value. For LPG's solution percentage/runtime performance, the data in Table 7 is a bit inconclusive. LPG and LPG+L seem to scale roughly similarly, which is also suggested by Figure 13. Across all the planners, the landmarks techniques bring about an improvement in the length of the found plans. The improvement is particularly strong for LPG, and there is only a single case (for FFv1.0, size $n = 38$) where the plans become a little longer using landmarks.

Our approximation methods can find various kinds of landmarks and orders in *Rovers*. The goal is a set of *communicated-data*-relations, where the data can be rock or soil analysis data, or images. For a rock or soil piece of data, the data can be gathered only by taking a sample at a specific waypoint *w*; if there is only a single rover *r* that is equipped for the right kind of analysis and that can navigate to *w* then we get *at(r w)* as a landmark.

22. Note that several orders of this kind may lead to cycles in the LGG, when the same truck has to change between two positions in both directions due to the transportation needs of different packages. We observed such cycles, and their removal, in our experiments.

| size | 30 | 32 | 34 | 36 | 38 | 40 |
|---|---|---|---|---|---|---|
| **%solved** | | | | | | |
| FFv1.0 | 50 | 45 | 30 | 15 | 25 | 5 |
| FFv1.0+L | 100 | 100 | 100 | 100 | 100 | 100 |
| FFv2.3 | 80 | 60 | 50 | 45 | 25 | 30 |
| FFv2.3+L | 100 | 100 | 100 | 100 | 100 | 100 |
| LPG | 60 | 70 | 55 | 40 | 35 | 10 |
| LPG+L | 70 | 25 | 45 | 45 | 40 | 10 |
| **time** | | | | | | |
| FFv1.0 | 181.21 | 115.81 | 217.87 | 90.57 | 167.01 | 182.38 |
| FFv1.0+L | 16.18 | 13.09 | 22.40 | 11.25 | 20.21 | 17.62 |
| FFv2.3 | 174.35 | 123.85 | 180.41 | 177.12 | 155.37 | 242.56 |
| FFv2.3+L | 19.71 | 17.76 | 28.49 | 27.73 | 28.34 | 36.68 |
| LPG | 153.42 | 75.40 | 202.18 | 175.65 | 176.25 | 247.36 |
| LPG+L | 108.41 | 90.24 | 183.14 | 174.32 | 177.60 | 218.69 |
| **length** | | | | | | |
| FFv1.0 | 206.10 | 130.78 | 229.50 | 88.67 | 145.80 | 186.00 |
| FFv1.0+L | 193.60 | 126.56 | 216.50 | 88.67 | 150.20 | 160.00 |
| FFv2.3 | 254.19 | 164.50 | 263.20 | 222.67 | 177.20 | 236.00 |
| FFv2.3+L | 245.62 | 159.42 | 262.70 | 219.44 | 170.60 | 226.50 |
| LPG | 231.75 | 119.75 | 281.25 | 229.50 | 190.00 | 203.50 |
| LPG+L | 218.38 | 115.00 | 246.88 | 211.00 | 172.57 | 178.00 |

Table 7: Experimental Results in *Rovers*. Time/plan length in each table entry is averaged over the respective instances solved by *both* of each pair "X" and "X+L" of planners. Size parameter is the number $n$ where the instances have $n$ rovers, $25 + n$ waypoints, $n$ cameras, 5 modes, and $5 + n$ objectives. 20 random instances per size.

For taking an image of an objective $o$ in a specific mode $m$, we need a camera $c$ that supports $m$, on a rover $r$ that can navigate to a point $w$ from which $o$ is visible. If there is only one appropriate combination of $c$, $r$, and $w$, then we get *at(r w)* and *calibrated(c)* as landmarks. In that case, if there is only one appropriate calibration target for $c$, we also get the respective rover position as a landmark. Reasonable orders arise, e.g., if there are several landmarks involving the position of the same rover. E.g., if a camera on a rover $r$ must be calibrated at a waypoint $w$, and be used to take an image at a waypoint $w'$, then we get a reasonable order *at(r w)* $\rightarrow_r$ *at(r w')*. Unsound orders can arise when the RPG wrongly approximates the structure of the road map. For example, an unsound order *at(r w)* $\rightarrow_{gn}$ *at(r w')* is extracted when *at(r w')* is a landmark candidate, in the RPG $r$ first reaches $w'$ from $w$, but really there are several locations from which $r$ can reach $w'$. The unsound $\rightarrow_{gn}$ orders can lead to unsound $\rightarrow_r$ orders. The unsound orders are not bad for the performance, though. When we used the "safe" approximation of $\rightarrow_n$ orders, the LGGs we got were non-trivial, namely basically the same as those with the RPG approximation, except for the unsound orders due to the wrong approximation of the road map. In terms of planner performance, compared to the RPG-approximated LGGs the sparser "safe" LGGs yielded longer runtimes (by about a factor 2), and also somewhat longer plans, for FFv1.0; for FFv2.3 runtime and plan length stayed roughly the same, except in the largest group of examples where the "safe"' LGGs yielded longer runtimes by a factor of 3; for LPG no significant changes were observable.

As for completeness of our approximations, we did not check this in full detail but it seems that our techniques do not overlook any valid orders in *Rovers*. (Except, as above in *Logistics*, reasonable orders $L \rightarrow_r L'$ where there is a path of greedy necesssary orders from $L$ to $L'$ anyway.) The only possible destructive interactions between landmarks are due to different positions of the same rover (which TIM detects as per fact inconsistency), or due to common delete effects of all achievers (as detected per condition 3 of Definition 6). The only possible reasons for (greedy) necessary orders, and for a fact being a landmark, lie in the form of the road maps of the individual rovers, and of their capabilities regarding taking rock/soil samples and images. All these structures are uncovered when using the RPG, i.e., for all facts $P$, the set of actions that can be used to achieve $P$ (for the first time) is a superset of the set of achievers of $P$ that appear (earliest) in the RPG.

In the presence of several rovers, where each rover can navigate to most of the waypoints as in the instances generated with Long and Fox's generator, only few landmarks and orders exist (because most of the time there are several alternative ways to do a certain job), making the decomposition imposed by our search control uninteresting. As said above, this is why we ran our experiments with instances featuring only a single rover.

### 6.10 Tyreworld

The *Tyreworld* domain was first designed by Stuart Russel (Russell & Norvig, 1995). Flat wheels on a car must be exchanged with spare tyres, which involves a number of different working steps. Problem size scales with the number of flat wheels. There are 13 operators to open/close the boot, to fetch/put away something from/into the boot, to loosen/tighten nuts, to jack up/jack down hubs, to do up/undo nuts, to put on/remove wheels, and to inflate wheels. To exchange a single wheel, one has to open the boot, fetch the spare wheel and the pump and the jack and the wrench, inflate the spare wheel, loosen the nuts on the right hub, jack up the hub, undo the nuts, remove the flat wheel, put on the spare wheel, do up the nuts, jack down the hub, and tighten the nuts. The tools must be put back into the boot at the end. There is only a single *Tyreworld* instance per size, and we used the generator provided on the FF homepage to generate examples with up to 30 flat wheels.[23] The results are very easy to interpret, so we do not include a detailed table here. FFv1.0 can solve tasks with up to 22 wheels within our 300 second time limit. FFv1.0+L comfortably scales up to the task with 30 wheels, which it solves in 14.28 seconds. For FFv2.3, which easily scales up to the task with 30 wheels (solved in 7.57 seconds) landmarks do not bring a runtime improvement (indeed, the runtimes become somewhat worse). For LPG, which also easily scales up to 30 wheels (14.55 seconds), the runtime behaviour is similar. As for plan length, this becomes somewhat longer for FFv1.0 and LPG when using landmarks; for FFv2.3, plan length remains the same.

Let us consider the (sub-)goals and orders in *Tyreworld* instances. Such instances have rather many top level goals. There are *in-boot* goals for all tools, and a *closed-boot* goal. For each flat wheel, we have an *on* goal for the respective spare wheel and hub, an *inflated* goal for the spare wheel, a *tight* goal for the nuts on the hub, and an *in-boot* goal for

23. It certainly sounds funny to encode a problem involving a car with 30 wheels all flat at the same time. But of course, what we are interested in here is the underlying *structure* of the problem, which does not depend on the names given to the predicates and operators.

the flat wheel. Reasonable orders between these goals tell us that, first, we should inflate the spare wheels, then mount these on their respective hubs, then tighten the nuts, then put the flat wheels and the tools into the boot, then close the boot. We do get some landmarks and orders beyond this, stating that at some point we must hold each spare wheel, and that at some point the hubs with flat wheels on must be made free. But this additional information is apparently not enough to make a significant difference over the Goal Agenda technique (which works with reasonable orders on top level goals). This explains the behaviour of FFv1.0 and FFv2.3: remember that these planners are basically the same except that FFv2.3 uses the Goal Agenda. Using landmarks, FFv1.0 becomes roughly as efficient as FFv2.3, at the cost of longer plans. For FFv2.3, in contrast, nothing much changes by using landmarks on top of the Goal Agenda, except an additional overhead for the landmarks computation. For LPG, it seems that this planner is very efficient in *Tyreworld* even without any goal ordering techniques. Our approximation algorithms are sound in *Tyreworld*, which is easy to see from the simple structure of the operators and initial states, which imply that all the approximated $\rightarrow_{gn}$ orders are, in fact, $\rightarrow_n$ orders. We did not check completeness in full detail, but could not find a case of a valid order that wasn't found by our approximation methods.

### 6.11 Unsound Orders

Briefly summarised, our observations regarding (the effect of) unsound orders in the RPG-approximated LGG are the following:

- *Blocksworld-arm* – no unsound orders generated.
- *Blocksworld-no-arm* – no unsound orders generated.
- *Depots* – no unsound orders generated.
- *Freecell* – unsound orders may be generated. With the sound approximation, one gets very few landmarks, but slightly more reliable behaviour.
- *Grid* – unsound orders may be generated. With the sound approximation, one gets very few landmarks, but somewhat better runtime behaviour. This is probably due to conflicts between the individual sub-tasks associated with the individual goal keys, arising when using the richer, potentially unsound, LGGs.
- *Logistics* – unsound orders may be generated, but are all removed prior to search since they refer to initial state facts. With the sound approximation, one gets very few landmarks, and a bad problem decomposition resulting in worse runtime and plan length behaviour.
- *Rovers* – unsound orders may be generated. With the sound approximation, one gets only slighty smaller sets of landmarks, resulting in similar, or somewhat worse, performance in our experiments.
- *Tyreworld* – no unsound orders generated.

## 7. Discussion

Various attempts have been made in the past to detect goal ordering constraints and use them for speeding up planning. All of these approaches differ considerably from our work. We list the most closely related ones. Irani and Cheng (1987) order a goal B after a goal A if, roughly speaking, A must be achieved before B can be achieved. The same authors later extended their notions to *sets* of goals (Cheng & Irani, 1989). In difference to our work, only top-level goals are considered. The same holds true for the work of Hüllem et al. (1999) who define a number of different ordering relations between top-level goals, which they approximate using (non-instantiated) operator graphs (Smith & Peot, 1993). A similar idea is explored in the PRECEDE system of McCluskey and Porteous (1997). They look at identifying necessary orders between top level goals but the focus of their approach is different: they develop a method for the construction of planning domain models and, within this, orders are identified *once* only for a given domain (rather than being problem instance specific) and are then exploited both for further validation of the domain model and also at planning time. Knoblock (1994) describes the ALPINE system, which *does* consider problem sub-goals to compute an "abstraction hierarchy" for the Prodigy planner (Fink & Veloso, 1994). In ALPINE, a sub-goal A is ordered before a sub-goal B if A must *possibly* be achieved first in order to achieve B. This is different from our work where orders of this kind are inserted when A must *necessarily* be achieved first in order to achieve B (also, the way these orders are used to structure the search is very different). Further, in difference to our approach, ALPINE does not consider the possible *negative* interactions between sub-goals, which we use to extract reasonable orders.

Koehler and Hoffmann (2000) provide the formal notion of reasonable orders between top-level goals, which they approximate and use for (potentially) splitting a planning task into several smaller sub-tasks. We extend this approach by the new dimension of necessary sub-goals, landmarks. By doing so, we obtain (potentially) more ordering information, e.g. in domains such as *Grid*, *Freecell*, *Logistics*, and *Rovers*, where the approach previously could do nothing to structure the search (because the achievement of one top level goal does not, or only very subtly, affect the achievement of another one). We provide a natural formal extension of reasonable orders to landmarks, clarify the computational complexity of our notions, and describe domain-independent and planner-independent algorithmic methods to approximate our definitions and to exploit the resulting ordering information. Our experiments show that our technology often speeds FF up significantly, sometimes dramatically, even when already using Koehler and Hoffmann's goal ordering techniques. A preliminary implementation of landmarks to control LPG shows that runtime gains are obtainable for this planner, too.

Various interesting research questions are open. For example, it is unclear to us why our landmarks technique often has less runtime impact on LPG than on FF; it ought to be investigated how our techniques can be better integrated with the LPG code. It might also be that, overall, there are alternative ways of using landmarks (e.g. as a plan initialisation technique rather than as a disjunctive search control framework) that work better for LPG. As another open topic, as said before our current technology removes, prior to planning, all cycles in the ordering relations. Instead, it could be investigated if these cycles can contain useful structural information, and how to exploit this information. Alternatively it could be investigated how several distinct points of truth of the same fact could be taken account of in the LGG, thereby avoiding the cycle problem altogether (or at least to a certain extent). It remains open how to avoid the performance loss that can arise if the planner frequently "switches" between conflicting parts of the LGG (c.f. our observations in *Grid*, Section 6.7). It also remains open how to lift our techniques to more powerful planning paradigms, in particular ADL, numeric, and temporal planning. As for ADL, lifting our definitions is probably straightforward, but extending the approximation techniques (e.g. the computation of "shared precondition facts" when preconditions are complex formulas) seems difficult. It probably makes more sense to look at an extension of STRIPS with conditional action effects only (planners such as FF (Hoffmann & Nebel, 2001) and IPP (Koehler et al., 1997) compile full-scale ADL down into such a format prior to planning). As for numeric planning, we believe that our definitions and approximation techniques can be naturally extended based on the "monotonicity" concept introduced with Metric-FF by Hoffmann (2003). Such techniques are likely to be useful in domains like *Settlers* (Long & Fox, 2003) where the numeric variables carry most of the information of what is needed to solve the task. In temporal planning, it seems our techniques could, in principle, be used with no extensions whatsoever (except adjusting them to the new syntax). It might be possible, however, to enrich the landmarks graph (the LGG) with information about the *time* that is at least needed to get from one landmark (a shared precondition, e.g.) to another one. Such additional information can probably play a useful role in the temporal planning process.

Finally, staying within the STRIPS setting, there are two interesting topics opened up by our preliminary handling of "lookahead necessary" orders. Currently we order $L \rightarrow_{ln} L'$ if $L'$ is a landmark, and there is an intermediate fact set $\{L_1, \ldots, L_m\}$ such that all achievers of $L'$ have (at least) one fact $L_j$ as precondition, and all achievers of any of the facts $L_j$ share the precondition $L$. Now:

1. The intermediate set $\{L_1, \ldots, L_m\}$ we use is a *disjunctive landmark* in the sense that $L_1 \vee \ldots \vee L_m$ must be true at some point on every solution plan. While currently we use $\{L_1, \ldots, L_m\}$ only to possibly find new landmarks $L$, disjunctive landmarks really are a generalisation of our approach. The landmarks we considered in this paper are the special case where the disjunctive landmarks are singleton sets. It remains to explore the general case. Interesting issues are to be resolved concerning the choice of disjunction candidates, and the definition of, e.g., reasonable orders.

2. While currently we look only one step ahead, in general one can look any number $k$ of steps ahead: find $k$ intermediate sets by backchaining over possible achievers, such that at the $k$th level there is a shared precondition. Obviously, doing so it would be critical to find good heuristics for selecting intermediate fact sets, and to trade the effort invested for the extended backchaining against the value of the obtained ordering information.

## Acknowledgments

The authors wish to thank Maria Fox and Derek Long for providing the TIM API. We are also very thankful to Alfonso Gerevini and Ivan Serina for their help with running/modifying the LPG system. We thank Malte Helmert for pointing out some NP-reductions. We finally wish to thank the anonymous reviewers for their comments, which were very helpful to improve the paper. This article is an extended and revised version of a paper (Porteous, Sebastia, & Hoffmann, 2001) that was published at ECP-01.

## Appendix A. Proofs

This appendix contains two kinds of proofs: first, statements about the computational complexity of deciding about our ordering relations; second, statements about our sufficient criteria for reasonable orders (i.e., implications from necessary orders and inconsistency to reasonable orders). Each kind of proofs is presented in one subsection.

### A.1 Computational Complexity

**Theorem 1** *Let* LANDMARK *denote the following problem: given a planning task $(A, I, G)$, and a fact $L$; is $L$ a landmark?*

*Deciding* LANDMARK *is PSPACE-complete.*

**Proof:** Hardness is proven by polynomially reducing the complement of PLANSAT– the decision problem of whether there exists a solution plan to a given arbitrary STRIPS planning task (Bylander, 1994) – to the problem of deciding LANDMARK. Given a planning task $(A, I, G)$, obtain the modified action set $A'$ by adding the three actions

$$\begin{array}{rrr} (pre, & add, & del) \\ (\{F\}, & I, & \{F\}), \\ (\{F\}, & \{L\}, & \{F\}), \\ (\{L\}, & G, & \emptyset) \end{array}$$

and obtain the modified initial and goal states by

$$I' := \{F\},\ G' := G$$

Here, $F$ and $L$ are new facts. In words, attach to $(A, I, G)$ an artificial way of by-passing the task. On that by-pass, $L$ must be added. $L$ is a landmark in the modified task iff $(A, I, G)$ is not solvable.

Membership follows because a fact is a landmark if and only if it is an initial- or goal-fact, or the task becomes unsolvable when ignoring all actions that add the fact. Deciding unsolvability – the complement of PLANSAT– is in PSPACE. □

**Theorem 2** *Let* NECESSARY-ORD *denote the following problem: given a planning task $(A, I, G)$, and two facts $L$ and $L'$; does $L \rightarrow_n L'$ hold?*

*Deciding* NECESSARY-ORD *is PSPACE-complete.*

**Proof:** Hardness is proven by polynomially reducing the complement of PLANSAT to the problem of deciding NECESSARY-ORD. Given a planning task $(A, I, G)$, obtain the modified action set $A'$ by adding the three actions

$$\begin{array}{rrr} (pre, & add, & del) \\ (\emptyset, & \{L\}, & \emptyset), \\ (\{L\}, & \{L'\}, & \emptyset) \\ (G, & \{L'\}, & \emptyset) \end{array}$$

and obtain the modified initial and goal states by

$$I' := I, \; G' := \{L, L'\}$$

Here, $L$ and $L'$ are new facts. The only chance of achieving $L'$ without having $L$ true a priori is when the original goal $G$ has been achieved. Thus, there is the necessary order $L \rightarrow_n L'$ if and only if the original task is unsolvable.

A non-deterministic algorithm that decides the complement of NECESSARY-ORD in polynomial space is the following. Iteratively guess actions, keeping in memory always only the current state and its predecessor. If $n$ is the total number of different facts in $(A, I, G)$, $2^n$ such guessing steps suffice to visit all reachable states (one can use a counter to keep track of the number of guessing steps already made). $L \rightarrow_n L'$ does not hold if and only if $L'$ is true initially, or there is a state that contains $L'$ such that the state's predecessor does not contain $L$. Finished with NPSPACE = co-PSPACE = PSPACE. □

Note that $L$ and $L'$ are landmarks in the hardness part to the proof of Theorem 2.

**Theorem 3** *Let* GREEDY-NECESSARY-ORD *denote the following problem: given a planning task $(A, I, G)$, and two facts $L$ and $L'$; does $L \rightarrow_{gn} L'$ hold?*

*Deciding* GREEDY-NECESSARY-ORD *is PSPACE-complete.*

**Proof:** Hardness is proven by exactly the same argument as in the proof to Theorem 2. For membership, one can use the same algorithm except that now one must also remember whether or not $L'$ has been true yet, and check only the predecessors of those states where $L'$ is true the first time. □

**Theorem 4** *Let* AFTERMATH *denote the following problem: given a planning task $(A, I, G)$, and two facts $L$ and $L'$; is $L'$ in the aftermath of $L$?*

*Deciding* AFTERMATH *is PSPACE-complete.*

**Proof:** Hardness is proven by reduction of the complement of PLANSAT. Given a planning task $(A, I, G)$, obtain the modified action set $A'$ by adding the actions

$$\begin{array}{rrr} (pre, & add, & del) \\ (\{F\}, & \{L'\}, & \{F\}), \\ (\{L'\}, & \{L\}, & \{L'\}), \\ (\{L\}, & I, & \{L\}), \\ (\{L\}, & \{L'\}, & \emptyset), \\ (\{L, L'\}, & G, & \emptyset) \end{array}$$

and obtain the modified initial and goal states by

$$I' := \{F\},\ G' := G$$

Here, $F$, $L$, and $L'$ are new facts. In words, what happens is this. The only state in $s \in S_{(L',\neg L)}$ is the one that results from applying the first new action to the initial state. One must then add $L$. Afterwards, there is the choice to either solve the original task, or achieve $L'$ and solve the task using the last of the new actions. There is a plan from $s$ on which $L'$ is not true at some point after $L$ if and only if the original task is solvable.

A non-deterministic algorithm that decides the complement of AFTERMATH in polynomial space is the following. Iteratively guess $2^n$ actions, where $n$ is the total number of different facts in $(A, I, G)$. $L'$ is not in the aftermath of $L$ if and only if there is an action sequence that contains a state $s \in S_{(L',\neg L)}$ and that achieves the goal, such that either $L$ is not true between $s$ and the first goal state, or $L'$ is not true between $L$ and the first goal state (checking $s \in S_{(L',\neg L)}$ and points where $L$, $L'$, or the goal are true the first time can be done by keeping appropriate flags). Finished with NPSPACE = co-PSPACE = PSPACE. □

**Theorem 5** *Let* REASONABLE-ORD *denote the following problem: given a planning task $(A, I, G)$, and two facts $L$ and $L'$ such that $L'$ is in the aftermath of $L$; does $L \rightarrow_r L'$ hold?*

*Deciding* REASONABLE-ORD *is PSPACE-complete.*

**Proof:** The hardness proof is exactly the same that Koehler and Hoffmann (Koehler & Hoffmann, 2000) use for proving PSPACE-hardness of deciding about reasonable orders, proceeding by a polynomial reduction of (the complement of) PLANSAT. Given an arbitrary planning task $(A, I, G)$, introduce new atoms $L$, $L'$, $F$ and $F'$ and obtain the modified action set $A'$ by adding the three actions

$$\begin{array}{rrr} (pre, & add, & del) \\ (\{F\}, & \{L', F'\}, & \{F\}), \\ (\{L', F'\}, & I, & \{F'\}), \\ (G, & \{L\}, & \emptyset) \end{array}$$

and modify the initial and goal states by

$$I' := \{F\},\ G' := G \cup \{L, L'\}$$

Here, $F$, $F'$, $L$, and $L'$ are new facts. $L$ and $L'$ are goals in the modified task so in particular $L'$ is in the aftermath of $L$. In the modified task, $S_{(L',\neg L)}$ contains the single state $\{L', F'\}$. From there, all plans that achieve $L$ delete $L'$ if and only if there is no such plan, i.e., if and only if the original task is unsolvable.

A non-deterministic algorithm that decides the complement of REASONABLE-ORD in polynomial space is the following. Iteratively guess $2^n$ actions, where $n$ is the total number of different facts in $(A, I, G)$, and check whether there is an action sequence that contains a state $s \in S_{(L',\neg L)}$, such that, after $s$, $L'$ does not get deleted before $L$ gets achieved (the necessary information about the sequence can be stored in flags). Finished with NPSPACE = co-PSPACE = PSPACE. □

### A.2 Sufficient Criteria

**Lemma 1** Given a planning task $(A, I, G)$, and two landmarks $L$ and $L'$. If either

1. $L' \in G$, or
2. there are landmarks $L = L_1, \ldots, L_{n+1}$, $n \geq 1$, $L_n \neq L'$, such that $L_i \rightarrow_{gn} L_{i+1}$ for $1 \leq i \leq n$, and $L' \rightarrow_{gn} L_{n+1}$, then $L'$ is in the aftermath of $L$.

**Proof:** The case where $L' \in G$ is trivial. We prove the other case. Let $s \in S_{(L',\neg L)}$, $s = Result(I, P)$, be a state where $L'$ was achieved strictly before $L$ on the action sequence $P$. As $L = L_1$ has never been true on the path $P$ from the initial state to $s$, neither has $L_2$ been true, or we have a contradiction to $L_1 \rightarrow_{gn} L_2$: $L_2$ is not true initially so it would have to be made true the first time at some point; immediately before this point, $L_1$ would need to be true. The same argument applies iteratively upwards to all pairs $L_i$ and $L_{i+1}$, implying that none of $L_1, \ldots, L_n$ have been true yet on the path to $s$. Likewise, $L_{n+1}$ can not have been true yet on the path to $s$.

Now, observe that, on any solution plan $P'$ from $s$, $L'$ is true one step before $L_{n+1}$ is true the first time. This simply follows from the greedy necessary order between the two facts. Say $P' = \langle a_1, \ldots, a_m \rangle$ is a solution plan from $s$, $G \subseteq Result(s, P')$. As $L_{n+1}$ is a landmark, and $L_{n+1}$ has not been true on the path to $s$, $P'$ fulfills $L_{n+1} \in Result(s, \langle a_1, \ldots, a_j \rangle)$ for some $1 \leq j \leq n$. Assume $j$ is the lowest such index. Then $L_{n+1}$ is made true the first time at the end of the path $P \circ \langle a_1, \ldots, a_j \rangle$. Therefore, with $L' \rightarrow_{gn} L_{n+1}$ we have $L' \in Result(s, \langle a_1, \ldots, a_{j-1} \rangle)$. So the first occurrence of $L_{n+1}$ is at a point $j$ while an occurrence of $L'$ is at $j - 1$.

We now arrive at our desired property by a simple backward chaining on the sequence $L_1, \ldots, L_n$. We know that $L_{n+1}$ is first true at point $j > 0$ on the action sequence $P' = \langle a_1, \ldots, a_m \rangle$. With $L_n \rightarrow_{gn} L_{n+1}$, we thus have that $L_n$ is true at $j - 1$. Because $L_n$ has not been true on the path to $s$, it first becomes true at some point $j'$ on the action sequence $P'$, where $j' < j$. We can apply the same argument downwards to $L_1$, giving us $L_1 = L \in Result(s, \langle a_1, \ldots, a_{j''} \rangle)$ for some $j'' < j$. This concludes the argument. □

**Theorem 6** Given a planning task $(A, I, G)$, and two landmarks $L$ and $L'$. If $L$ interferes with $L'$, and either

1. $L' \in G$, or
2. there are landmarks $L = L_1, \ldots, L_{n+1}$, $n \geq 1$, $L_n \neq L'$, such that $L_i \rightarrow_{gn} L_{i+1}$ for $1 \leq i \leq n$, and $L' \rightarrow_{gn} L_{n+1}$, then there is a reasonable order between $L$ and $L'$, $L \rightarrow_r L'$.

**Proof:** By Lemma 1, we know that $L'$ is in the aftermath of $L$. Say $s \in S_{(L',\neg L)}$. We need to prove that any action sequence achieving $L$ from $s$ deletes $L'$. We need to cover four reasons for interference.

1. $L$ is inconsistent with $L'$: with inconsistency, from *any* reachable state were $L'$ is true we must delete $L'$ in order to achieve $L$.
2. there is a fact $x \in \bigcap_{a \in A, L \in add(a)} add(a)$ such that $x$ is inconsistent with $L'$: $x$ will necessarily be true in the state after we achieved $L$, so $L'$ must be deleted by some action on the way to that state.
3. $L' \in \bigcap_{a \in A, L \in add(a)} del(a)$: obvious, the action adding $L$ will delete $L'$.
4. Say there is an $x$ that is inconsistent with $L'$, and $x \rightarrow_{gn} L$. We know that $s$ is reachable, $s = Result(I, P)$, $L' \in s$, and $L$ is not true at any point on the path to $s$. As $L' \in s$, we have $x \notin s$ with inconsistency. Say $P' = \langle a_1, \ldots, a_n \rangle \in A^*$ is an action sequence such that $L \in Result(s, P')$, i.e., $L \in Result(s, \langle a_1, \ldots, a_i \rangle)$ such that $i$ is minimal. Because $L$ is achieved the first time on the sequence $P \circ \langle a_1, \ldots, a_i \rangle$, we can apply the greedy necessary order $x \rightarrow_{gn} L$, so we know that $x \in Result(I, P \circ \langle a_1, \ldots, a_{i-1} \rangle)$ ($i \geq 1$ because $L \notin s$). With inconsistency, $L' \notin Result(I, P \circ \langle a_1, \ldots, a_{i-1} \rangle)$, so $i - 1 > 0$ and $L' \in del(a_j)$ for some $1 \leq j \leq i - 1$.

□

**Lemma 2** Given a planning task $(A, I, G)$, a set $O$ of reasonable ordering constraints, and two landmarks $L$ and $L'$. If either

1. $L' \in G$, or
2. there are landmarks $L = L_1, \ldots, L_{n+1}$, $n \geq 1$, $L_n \neq L'$, such that $L_i \rightarrow_{gn} L_{i+1}$ or $L_i \rightarrow_r L_{i+1} \in O$ for $1 \leq i \leq n$, and $L' \rightarrow_{gn} L_{n+1}$, then $L'$ is in the obedient aftermath of $L$.

**Proof:** The case where $L' \in G$ is trivial. We prove the other case. Let $s \in S^O_{(L',\neg L)}$ be a state where $L'$ has been achieved strictly before $L$; let $P$ be an obedient path from the initial state to $s$ on which $L$ has not yet been true, and $L'$ has just been achieved. As $L = L_1$ has never been true on the path, neither has $L_2$ been true, or we have a contradiction to the order between $L_1$ and $L_2$. The same argument applies iteratively upwards to all pairs $L_i$ and $L_{i+1}$, implying that none of $L_1, \ldots, L_n$ have been true yet on the path to $s$. Likewise, $L_{n+1}$ can not have been true yet on the path to $s$ if the second case of the prerequisite holds. In particular, none of these facts is contained in $s$ itself.

Now, observe that, on any solution plan $P'$ from $s$, $L'$ is true one step before $L_{n+1}$ is true the first time. This simply follows from the greedy necessary order between the two facts. Say $P' = \langle a_1, \ldots, a_m \rangle$ is a solution plan from $s$, $G \subseteq \mathit{Result}(s, P')$, such that $P \circ P'$ is obedient. As $L_{n+1}$ is a landmark, and $L_{n+1}$ has not been true on the path to $s$, $P'$ fulfills $L_{n+1} \in \mathit{Result}(s, \langle a_1, \ldots, a_j \rangle)$ for some $1 \leq j \leq n$. Assume $j$ is the lowest such index. Then $L_{n+1}$ is made true the first time at the end of the path $P \circ \langle a_1, \ldots, a_j \rangle$. Therefore, with $L' \rightarrow_{gn} L_{n+1}$ we have $L' \in \mathit{Result}(s, \langle a_1, \ldots, a_{j-1} \rangle)$. So the first occurrence of $L_{n+1}$ is at a point $j$ while an occurrence of $L'$ is at $j-1$.

We now arrive at our desired property by a simple backward chaining on the sequence $L_1, \ldots, L_n$. We know that $L_{n+1}$ is first true at point $j > 0$ on the action sequence $P' = \langle a_1, \ldots, a_m \rangle$. If $L_n \rightarrow_{gn} L_{n+1}$, we thus have that $L_n$ is true at $j-1$. If $L_n \rightarrow_r L_{n+1} \in O$, we have that $L_n$ is true at some $0 < j' < j$, because $P \circ P'$ is obedient and $L_n$ has not been true on the path to $s$. In both cases, we know that $L_n$ becomes true the first time at some point $j'$ on the action sequence $P'$, where $j' < j$. We can apply the same argument downwards to $L_1$, giving us $L_1 = L \in \mathit{Result}(s, \langle a_1, \ldots, a_{j''} \rangle)$ for some $j'' < j$. This concludes the argument. □

**Theorem 7** Given a planning task $(A, I, G)$, a set $O$ of reasonable ordering constraints, and two landmarks $L$ and $L'$. If $L$ interferes with $L'$, and either

1. $L' \in G$, or

2. there are landmarks $L = L_1, \ldots, L_{n+1}$, $n \geq 1$, $L_n \neq L'$, such that $L_i \rightarrow_{gn} L_{i+1}$ or $L_i \rightarrow_r L_{i+1} \in O$ for $1 \leq i \leq n$, and $L' \rightarrow_{gn} L_{n+1}$, then there is an obedient reasonable order between $L$ and $L'$, $L \leq_r^O L'$.

**Proof:** By Lemma 2, we know that $L'$ is in the obedient aftermath of $L$. Say $s \in S^O_{(L', \neg L)}$. We need to prove that any action sequence achieving $L$ from $s$ deletes $L'$. For the four reasons for interference, this can be proven with exactly the same arguments as in the proof to Theorem 6. □

## Appendix B. Runtime Distribution Graphs

This appendix provides supplementary material to the experimental data presented in Section 6. In Figures 11, 12, and 13, we provide runtime distribution graphs for all the testing domains (except *Tyreworld* where, as described in Section 6.10, the results are very easy to interpret). We plot the number of solved instances against the elapsed runtime. We do not make a distinction between the different sizes of the instances in the example suites, i.e., the numbers of solved instances shown refer to the entire set of example instances in the respective domain. The graphs are (only) intended to foster understandability of the data presented in Section 6, and should be self-explanatory given the discussion in that section.

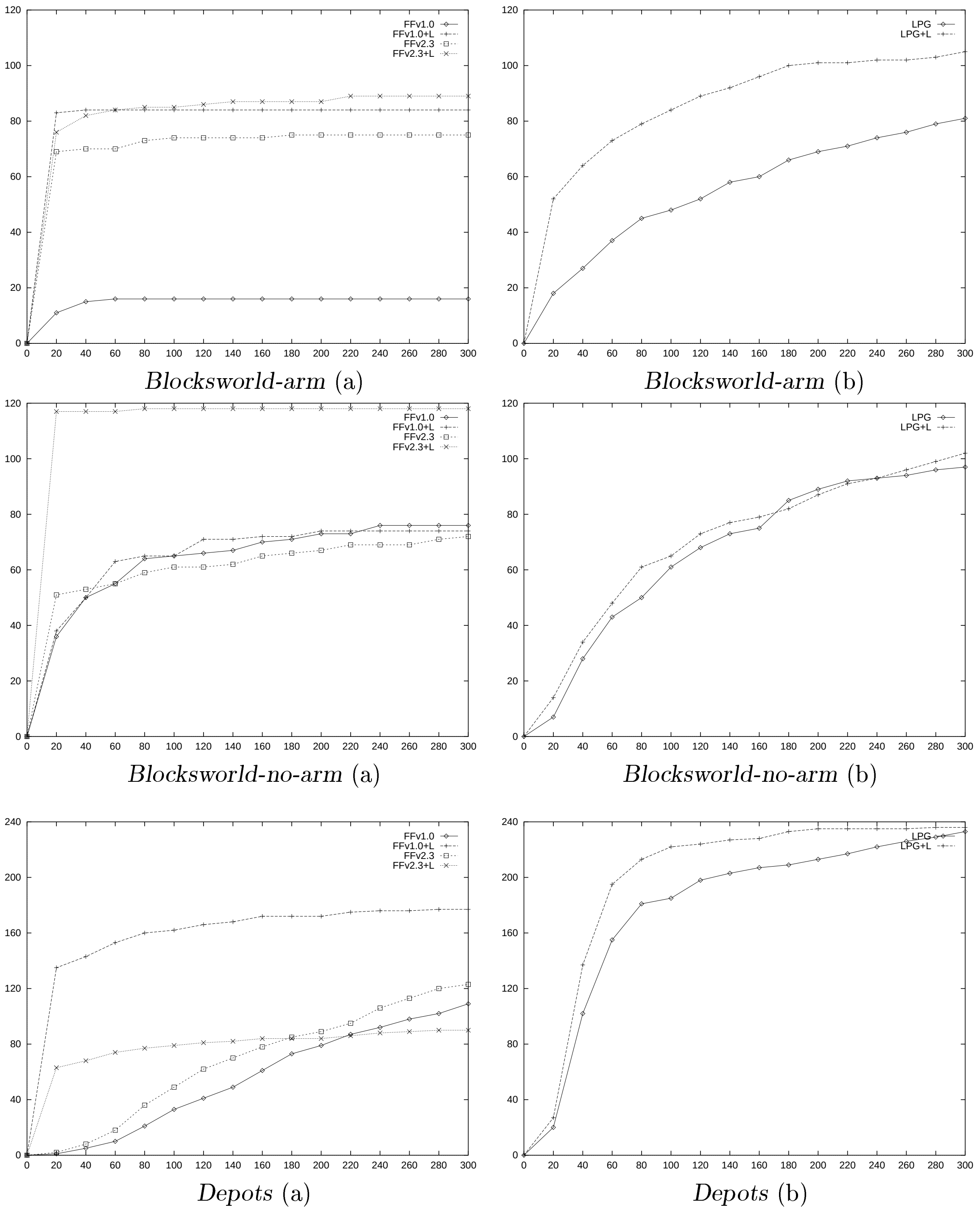

Figure 11: Number of solved instances ($y$-axis) plotted against seconds runtime ($x$-axis) for (a) FFv1.0 and FFv2.3, as well as for (b) LPG. Curves shown for each planner without landmarks control ("X"), and for each planner with landmarks control ("X+L"), in the *Blocksworld-arm*, *Blocksworld-no-arm*, and *Depots* example suite from Section 6.

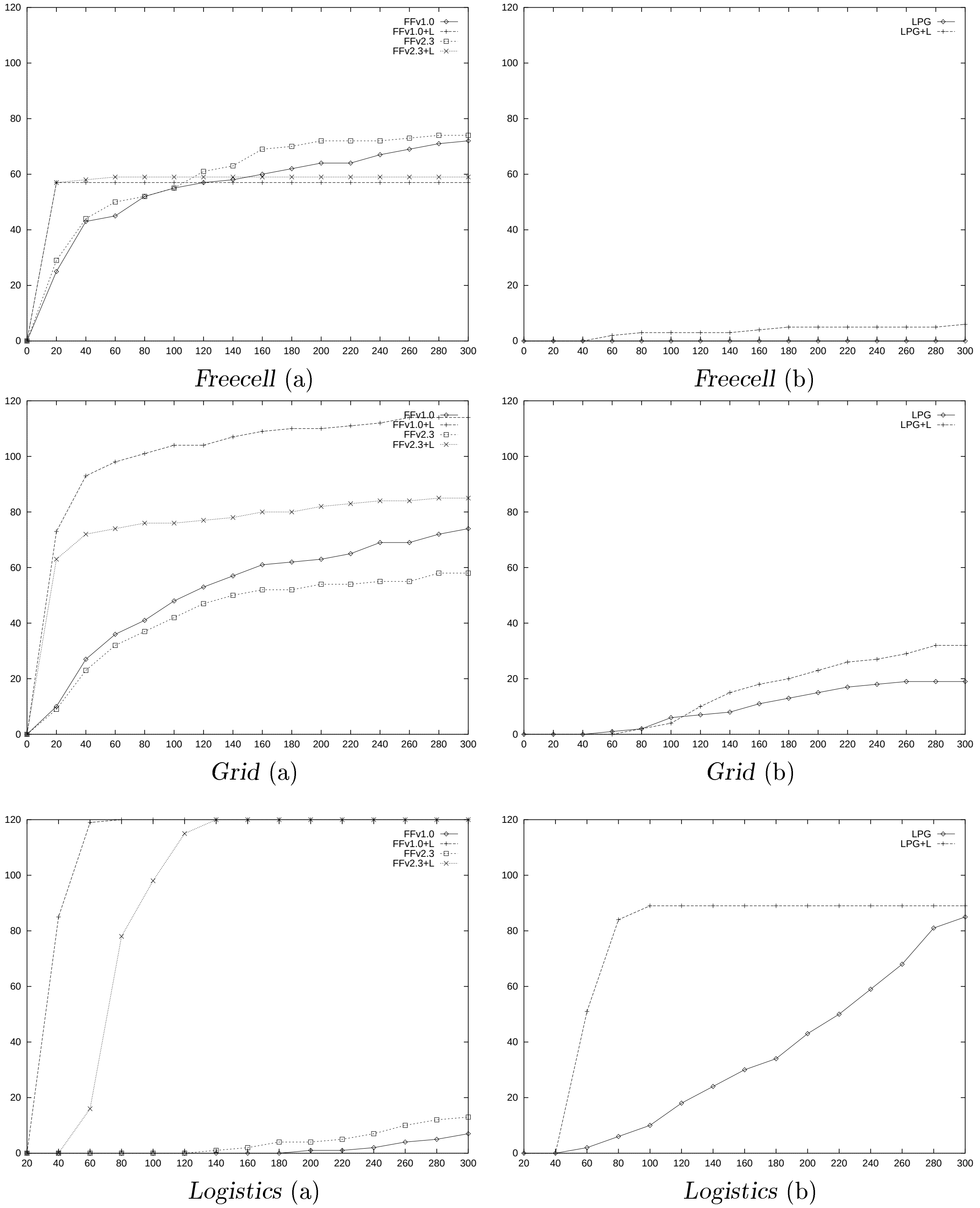

Figure 12: Number of solved instances ($y$-axis) plotted against seconds runtime ($x$-axis) for (a) FFv1.0 and FFv2.3, as well as for (b) LPG. Curves shown for each planner without landmarks control ("X"), and for each planner with landmarks control ("X+L"), in the *Freecell*, *Grid*, and *Logistics* example suites from Section 6.

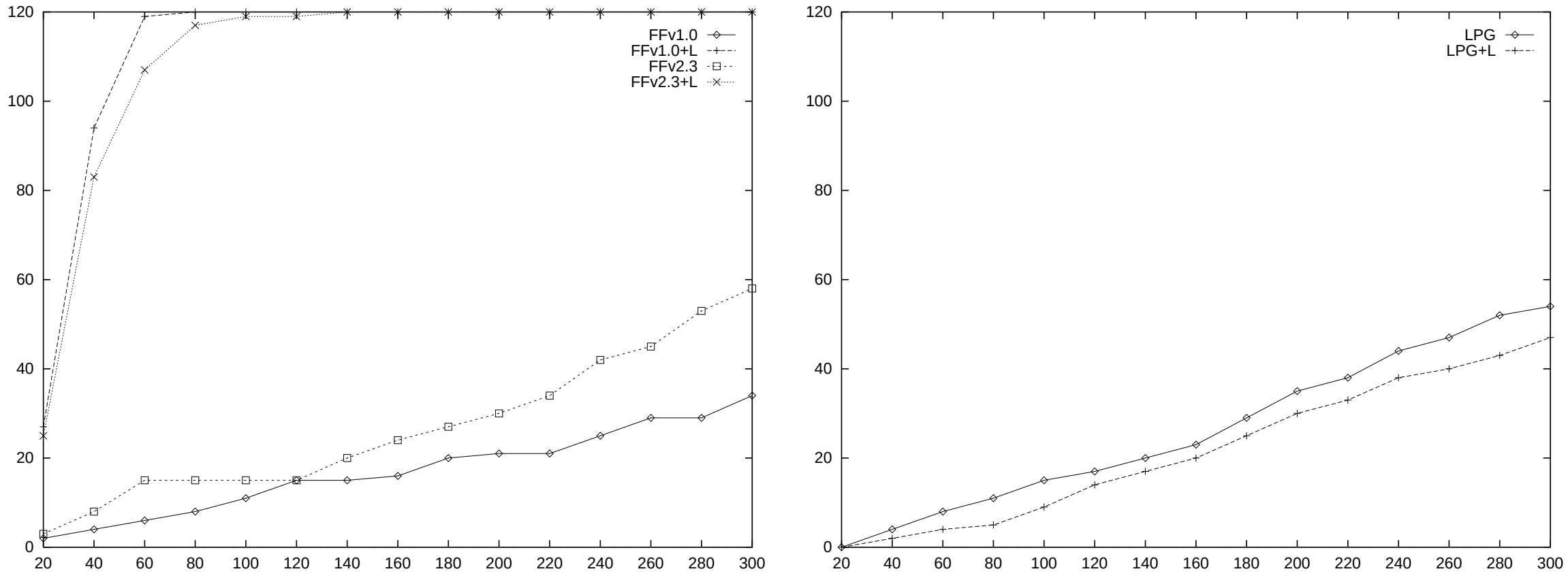

*Rovers* (a) *Rovers* (b)

Figure 13: Number of solved instances ($y$-axis) plotted against seconds runtime ($x$-axis) for (a) FFv1.0 and FFv2.3, as well as for (b) LPG. Curves shown for each planner without landmarks control (“X”), and for each planner with landmarks control (“X+L”), in the *Rovers* example suites from Section 6.

## Appendix C. Experimental LPG+L Implementation

In this appendix we discuss our experimental implementation of landmarks control around LPG. The implementation is preliminary in that it does not integrate landmarks control directly with LPG's code, but rather implements a simple control loop around an LPG executable. This causes a large runtime overhead due to the repetition of LPG's pre-processing machinery that computes fact (and action) inconsistencies. Our approach is to simply ignore this unnecessary overhead. This way, our experimental code produces the same results as a direct implementation that would do the pre-process only once – *provided LPG's pre-process computes the same information throughout the control loop.* We show that the latter is indeed the case in all our 8 testing domains except in *Freecell*. As explained in Section 6 already, because of these results in *Freecell* we measure total running time of our implementation, while in the other 7 domains we count the runtime for only the first one of LPG's inconsistency pre-processes.

Let us first discuss LPG's overall architecture. As far as relevant for us here, it proceeds as follows:

1. Perform a pre-process that detects inconsistent facts and actions.

2. Build a planning graph where the mutex reasoning is replaced by inserting the pre-computed inconsistencies.

3. Perform a local search in a space of sub-graphs of the resulting planning graph.

We need a little more detail about steps 1 and 2. We summarise the algorithms described by Gerevini et al. (2003). Step 1 first computes inconsistent pairs of facts. The algorithm keeps track of sets $F$ and $A$ of reached facts and actions, and of a set $M$ of potential inconsistencies (persistent mutex relations, in LPG's terminology). $F$ is initialised with the initial state of the task that LPG is run on, $A$ and $M$ are initialised with the empty set. Then a fixpoint procedure iterates until there are no more changes to $F$ and $M$. In each iteration, all actions are considered whose preconditions are in $F$ without an inconsistency in $M$. Such actions are used to identify new potential inconsistencies (e.g., between the action's add and delete effects). Potential inconsistencies are removed when there are actions in $A$ that either add both facts, or add one fact without deleting the other. When the fixpoint is reached, the fact pairs in $M$ are proved to be inconsistent. The inconsistent actions are then computed as those that either interfere (delete each others add effects or preconditions), or have competing needs (inconsistent preconditions). Step 2 in LPG's architecture then builds a planning graph without mutex reasoning. Starting from the initial state, fact layers $F_i$ and action layers $A_i$ are built alternatingly, where $A_i$ contains all actions whose preconditions are in $F_i$, and $F_{i+1}$ contains all add effects of the actions in $A_i$. The process stops with a layer where no new facts come in, and the mutex pairs in the resulting graph are the inconsistent pairs computed by step 1.

As said, our experimental implementation repeatedly calls an LPG executable inside the landmarks control. So steps 1 and 2 are done over again, producing a large runtime overhead. In an implementation integrating the landmarks control directly into LPG's code, one could avoid much of this overhead by using an overall architecture that computes the inconsistencies only once, and (only) does steps 2 and 3 of LPG's architecture for every sub-task in the control loop. We henceforth refer to such a hypothetical implementation as *direct-LPG+L*, and to our experimental implementation as *LPG+L*. The idea is to simulate direct-LPG+L's behaviour by running LPG+L, and counting the runtime taken by LPG in step 1 only for the first iteration of the control loop.[24] The subtlety to be taken care of is that the result of step 1 depends on the start (initial) state that LPG is run on. Remember that the search control loop in LPG+L calls LPG on a sequence of consecutive start states $s$, where the first $s$ is the initial state, and each successor start state $s'$ results from the previous start state $s$ by applying the plan that LPG found for the last sub-task. Now, different found inconsistencies (for different start states) in step 1 can result in different planning graphs at the end of step 2, and different planning graphs can result in different search behaviour in step 3. So we can only safely simulate direct-LPG+L in cases where the planning graph resulting from step 2 is always the same in LPG+L as it would be in direct-LPG+L. We found that the latter is indeed the case in 7 of our 8 testing domains. The reason for this is that state reachability is (largely, in some cases) invertible in these domains.

Consider a state $s'$ that is reachable from a state $s$. Denote by $F(s)$, $A(s)$ and $M(s)$ the sets $F$, $A$, and $M$ at the end of LPG's fact inconsistency fixpoint computation when started with initial state $s$, similarly for $s'$. Then $F(s') \subseteq F(s)$, $A(s') \subseteq A(s)$, and $M(s') \supseteq M(s) \cap (F(s') \times F(s'))$ hold. The reason is simply that everything that is reachable from $s'$ is also reachable from $s$. The fixpoint process from $s$ will eventually "execute" the action sequence that leads from $s$ to $s'$, i.e., include these actions into the set $A$. At this point,

24. The LPG implementation interleaves the computations for step 1 and step 2, which is why the individual runtimes can not be separated, and our LPG+L implementation in fact ignores the overhead for both step 1 *and* step 2 except in the first iteration of the control loop. The time taken for step 2 is typically not significant. See also the discussion at the end of this appendix.

$s'$ will be contained in the set $F$, with no potential inconsistencies. It follows that every fact or action that the process reaches when started in $s'$ will also be reached when started in $s$. It also follows that every potential inconsistency the process removes from $M$ when started in $s'$ will also be removed when started in $s$. This proves the claim. (Actually, $M(s') \supseteq M(s) \cap (F(s') \times F(s'))$ already follows because every fact pair that is inconsistent below $s$ must also be inconsistent below $s'$. We include the longer proof sketch to provide a better background for the proof arguments below.)

If LPG's step 1 finds the same fact inconsistencies for two states, then it trivially follows that the action inconsistencies identified will also be the same. Thus, with the above, if states $s$ and $s'$ are both reachable from each other, then the outcome of step 1 is the same for both states. This immediately shows that in invertible domains – domains where to each action $a$ there is an inverse action $\overline{a}$ that undoes exactly $a$'s effects – the outcome of LPG's step 1 remains the same throughout our landmarks control loop. *Blocksworld-arm*, *Blocksworld-no-arm*, *Depots*, and *Logistics* are all invertible.

The other three of our experimental domains where we can safely ignore the overhead caused by step 1 repetition are *Grid*, *Rovers*, and *Tyreworld*. For each of these domains, we show that the outcome of step 2 in direct-LPG+L is the same as the outcome of step 2 in LPG+L, for every iteration of the landmarks control loop. Let us start with the *Tyreworld*. Consider the information computed by LPG's steps 1 and 2 in states $s$ and $s'$, where $s'$ is reachable from $s$. *Tyreworld* is not invertible only because one can not "deflate" spare wheels that one has already inflated. From $s'$ one can get to a state $s''$ that subsumes $s$ except for, possibly, a set of *not-inflated(?spare-wheel)* facts. Such facts serve only as precondition for the actions that inflate *?spare-wheel*, which actions only achieve *inflated(?spare-wheel)*. These *inflated(?spare-wheel)* facts are already contained in $s'$. So the fact inconsistency fixpoint process can reach everything from $s'$ that it can reach from $s$, except the inflate *?spare-wheel* actions. It follows that the only difference of the outcome of step 1 in $s'$ compared to its outcome in $s$ is that from $s$ there can be inconsistencies involving the inflate *?spare-wheel* actions. But when step 2 is done in $s'$, these actions are never reached anyway as their *not-inflated(?spare-wheel)* preconditions are not reachable. So steps 2 in both LPG+L and direct-LPG+L end up with the same planning graphs for the state $s'$.

A similar argument as in *Tyreworld* applies to states $s$ and $s'$ in *Grid* tasks. The only thing that can not be undone in *Grid* tasks is the opening of locked locations. From $s'$ one can get to a state $s''$ that subsumes $s$ except, possibly, a set of *locked(?location)* facts. Such facts serve only as precondition for actions that unlock *?location*, which actions only achieve *unlocked(?location)*. These *unlocked(?location)* facts are already contained in $s'$, and the only difference of the outcome of step 1 in $s'$ compared to its outcome in $s$ is that from $s$ there can be inconsistencies involving actions that unlock *?location*. When direct-LPG+L does step 2 in $s'$, these actions are never reached anyway.

Finally, let's consider the *Rovers* domain. When reaching a state $s'$ from a state $s$ in *Rovers*, one can in general not get back to $s$, because once a soil or rock sample has been taken, it can not be put back to the waypoint it's been taken from. That is, from $s'$ one can get to a state $s''$ that subsumes $s$ except, possibly, a set of *at-soil/rock-sample* facts.[25] This, in turn, can only affect the reachability of *sample-soil/rock* actions, *have-soil/rock-sample* facts, *full-storage-space* facts, *communicate-soil/rock-data* actions, and *communicated-rock/soil-data* facts. So the only difference of the outcome of step 1 in $s'$ compared to its outcome in $s$ is that from $s$ there can be inconsistencies involving these facts and actions. But when direct-LPG+L does step 2 in $s'$, these facts and actions are never reached anyway.

We finally remark the following. The LPG implementation interleaves the computations for step 1 and step 2, and LPG outputs the runtime taken by step 1 *together with* step 2. So our LPG+L implementation in fact ignores the overhead for both step 1 *and* step 2 except in the first iteration of the control loop. In personal communication, Alfonso Gerevini and Ivan Serina informed us that step 2 typically takes only a small fraction (around 5%) of the time spent in pre-processing – which seems reasonable considering that step 2 is basically the same as building a relaxed planning graph, i.e., a single heuristic evaluation in, e.g., FF. Nevertheless, we reiterate that our LPG+L implementation is preliminary, and that the runtime results should be treated with care. In a direct implementation of landmarks in LPG, step 2 would have to be done for every iteration of the control loop, which runtime overhead we could not count due to the above implementational difficulties. On the other hand, in a direct implementation one could exploit various ideas regarding the computation of inconsistencies for different start states. Alfonso Gerevini suggested that, e.g., one could modify the inconsistency computation to only take account of the *changes* in a new state. One could also derive state invariants, like DISCOPLAN (Gerevini & Schubert, 2000) does, that are computed only once (independently of the initial state), and then pruned accordingly every time a new start state comes up (pruning amounts to removing those invariants that do not hold given a particular initial state). Finally, one could try to replace LPG's inconsistency reasoning with the TIM inconsistency function, thereby avoiding step 1 altogether. As we said before, exploring such possibilities is a topic for future work.

25. Taking an image deletes the calibration of the camera; but one can always re-calibrate the camera with the same calibration target as was used before – no camera is calibrated initially.